\documentclass[10pt, a4paper]{article}
\usepackage[utf8]{inputenc}

\usepackage{graphicx}
\usepackage{amsmath,amssymb,amsfonts}
\usepackage{textcomp,marvosym}
\usepackage{changepage}
\usepackage{rotating}
\usepackage{booktabs}
\usepackage{longtable}
\usepackage{array}
\usepackage{multirow}
\usepackage{makecell}
\usepackage{float}
\usepackage{pdflscape}
\usepackage{lmodern}
\usepackage{ragged2e}
\usepackage{subcaption}
\usepackage{adjustbox}
\usepackage{caption}
\usepackage[table]{xcolor}
\definecolor{lightgrayrow}{RGB}{245,245,245}
\usepackage{makecell}
\usepackage{amssymb}
\usepackage{colortbl}
\usepackage{pifont}
\usepackage{url}
\usepackage[margin=1in]{geometry}
\usepackage{tikz}
\usetikzlibrary{fit, backgrounds}
\usetikzlibrary{positioning,calc,arrows.meta,shapes.geometric}

\usepackage{pgfplots}
\pgfplotsset{compat=1.18}
\usepgfplotslibrary{groupplots}

\usepackage[numbers,sort&compress]{natbib}

\usepackage{hyperref}
\usepackage{cleveref}
\usepackage{authblk}

\usepackage[margin=1in]{geometry}
\usepackage[utf8]{inputenc}
\usepackage[T1]{fontenc}
\usepackage{hyperref}
\usepackage{amsmath, amssymb}

\newcommand\blfootnote[1]{%
  \begingroup
  \renewcommand\thefootnote{}\footnote{#1}%
  \addtocounter{footnote}{-1}%
  \endgroup
}

\graphicspath{{./}}
\usepackage{lineno}
\modulolinenumbers[1]

\newcolumntype{P}[1]{>{\raggedright\arraybackslash}p{#1}}

\hypersetup{
    linktocpage=true,
    colorlinks=true,
    citecolor=blue,
    urlcolor=blue,
    linkcolor=blue,
    breaklinks=true
}

\title{Robotic Strawberry Harvesting with Robust Vision and Deep Reinforcement Learning based Sim-to-Real Control}

\author{
    Al Bashir$^{a, b, \dagger}$, Shao-Yang Chang$^{c, \dagger}$, Partho Ghose$^{a, b}$, Prem Raj$^{a, b}$, Chen-Kang Huang$^{c}$ and Azlan Zahid$^{a, b, *}$
}

\date{}

\begin{document}


\maketitle
\begin{center}
    \vspace{-2em}
    {\small
    $^{a}$ Department of Biological \& Agricultural Engineering, Texas A\&M University, College Station, TX 77843, USA\\
    $^{b}$ Texas A\&M AgriLife Research, Texas A\&M University System, Dallas, TX 75252, USA\\
    $^{c}$Department of Biomechatronics Engineering, National Taiwan University, 10617, Taipei City, Taiwan
    }
\end{center}
\vspace{1em}

\blfootnote{$^\dagger$ These authors contributed equally}
\blfootnote{$^*$ Corresponding author}
\blfootnote{\texttt{albashir@tamu.edu} (A. Bashir); \texttt{r13631040@ntu.edu.tw} (S. Chang); \texttt{partho.ghose@tamu.edu} (P. Ghose); \texttt{prem.raj@tamu.edu} (P. Raj); \texttt{ckhuang94530@ntu.edu.tw} (C. Huang); \texttt{azlan.zahid@tamu.edu} (A. Zahid)}
\blfootnote{ORCID(s): 0000-0003-2168-2840 (A. Bashir); 0000-0002-5840-2008 (P. Ghose); 0000-0001-6202-8680 (A. Zahid)}

\begin{abstract}
This study presents a closed-loop robotic strawberry harvesting system that combines a robust vision module, simulation-trained deep reinforcement learning (DRL) control, and ROS-based real-robot execution. For perception, we propose HRAttnEdge-YOLO26-seg, a modified YOLO26-seg architecture that incorporates a high-resolution P2 branch, segmentation-path attention, and edge-supervised prototype learning to improve instance segmentation in cluttered scenes. For control, we train a target-conditioned Proximal Policy Optimization (PPO) policy in Isaac Lab to produce smooth joint-position commands for a UR10e manipulator and deploy it on a UR10e robot for target-fruit reaching and harvesting. This simulation-based approach reduces hardware dependency, lowers development cost, and allows scalable policy training without exhaustive physical trials before real deployment. The proposed vision model demonstrated the highest overall performance among the evaluated methods. On both self-collected and public datasets, the model showed a 10 to 14\% improvement in segmentation performance. In controlled in-house tests, the PPO controller produced stable and dynamically smoother motion than a inverse kinematics (IK)-based MoveIt baseline. In greenhouse trials, the proposed integrated system harvested 281 strawberries, achieving 96.6\% reaching success, 91.3\% grasp-and-pull success, and 84.3\% overall harvesting success. These results illustrate that task-specific perception combined with simulation-trained PPO can serve as a practical and resource-efficient alternative to conventional planner-dependent reaching in manipulation, enabling reliable closed-loop robotic harvesting in complex agricultural environments.

\vspace{1em}
\noindent\textbf{Keywords:} Instance segmentation, Robot vision, PPO, Trajectory control, Isaac Lab, Motion planning, Agricultural robotics.
\end{abstract}
\begin{figure}[t]
\centering
\includegraphics[width=.80\textwidth]{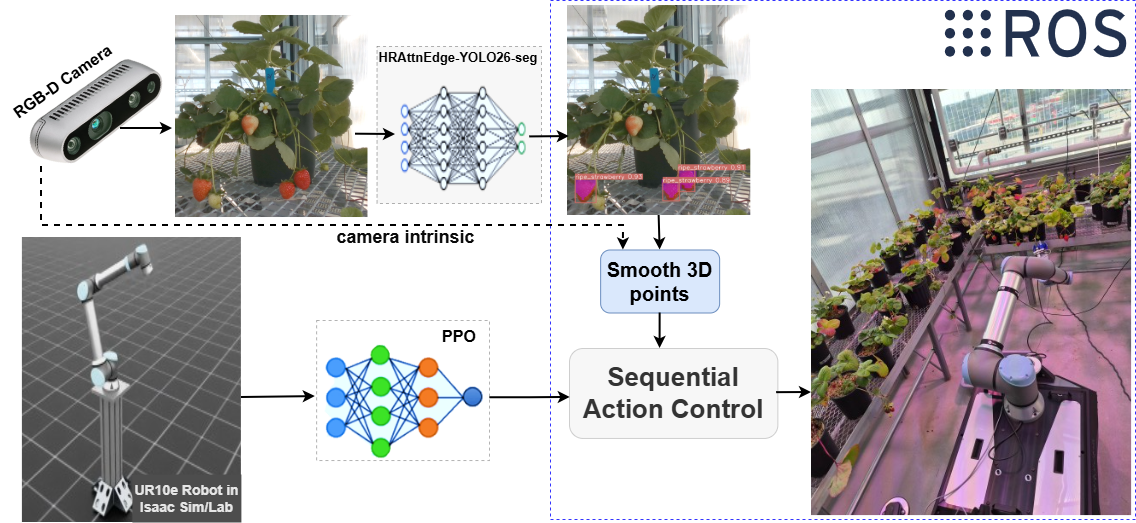}
\caption{Overview of the closed-loop robotic strawberry harvesting system. An RGB-D camera acquires greenhouse fruit scenes, and the proposed \emph{HRAttnEdge-YOLO26-seg} model segments target strawberries. 
Smoothed 3D target points are derived from the center point of the segmented mask, corresponding depth, and camera intrinsics. In parallel, a PPO policy trained in Isaac Sim/Lab generates reaching motions of the UR10e robot towards the target. All processes, including perception, target generation, robot motion, and gripper-level harvesting actions, are integrated and executed through ROS.}  
\label{fig:overview}
\end{figure}

\section{Introduction}
Harvesting is one of the most labor-intensive and cost-sensitive operations in agricultural production~\cite{Stemmler2023}. The problem is particularly acute in high-value specialty crops, such as strawberries, where fruit must be selected individually, handled carefully, and removed within a narrow window of maturity~\cite{schimmelpfennig2018crop, pankova2020economic}. In controlled-environment agriculture (CEA) and indoor farming systems, dense planting layouts, repeated picking cycles, and strict fruit-quality requirements further intensify these demands~\cite{getahun2024application}. At the same time, persistent labor shortages~\cite{win2025labor} and increasing labor costs~\cite{charlton2019can} continue to accelerate the demand for intelligent robotic harvesting systems~\cite{pandey2026robotics}. Despite recent progress, achieving reliable robotic harvesting remains challenging because the system must simultaneously solve three tightly coupled problems: accurate fruit perception in cluttered scenes, stable 3D target estimation for manipulation, and efficient motion generation under uncertainty and workspace constraints~\cite{wang2025research, rajendran2024towards}.

Prior works have explored a wide range of perception and robotic control strategies \cite{WU2026111409, arad2020development}. For perception, while geometry and point-cloud-based approaches improve spatial reasoning, they typically introduce additional sensing, calibration, and computational overhead that limits real-time deployment~\cite{noda2025robot, catala2024harnessing}. Consequently, deep learning-based object detection and segmentation frameworks have been extensively adopted for fruit localization~\cite{HUANG2026111616, jia2020detection}. On the manipulation side, most harvesting systems rely on inverse-kinematics (IK)~\cite{HUAPENG2001321}, motion-planning, or visual-servoing-based controllers~\cite{al2024enhancing, calzada2025review}. For example, Huang Z. et al.~\cite{HUANG2026111616} and Wu S. et al. ~\cite{WU2026111409} consecutively developed robotic cucumber and camellia oleifera fruit-harvesting systems using a Fuzzy self-attention enhanced YOLOv8n~\cite{yolov8_ultralytics} and a modified YOLO11~\cite{yolo11_ultralytics} detector models for perception. However, the control module remained dependent on Hybrid Visual Servo Control (HVSC)~\cite{malis20022}, which is computationally complex~\cite{malis20022}, sensitive to calibration and environmental noise~\cite{chaumette2006visual, 1321162}, and vulnerable to singularities that can induce unstable motion~\cite{levine2018learning}. On the other hand, Arad B. et al. \cite{arad2020development} and Bhat K. M. et al. \cite{bhat2026vision} developed IK-based motion control for robotic harvesting of peppers. However, their dependence on general-purpose perception techniques, such as color thresholding and a pre-trained YOLO11 model, restricts the effectiveness of downstream robotic operations. Because general-purpose perception often yields unstable masks and inaccurate boundaries in highly cluttered agricultural environments, resulting in unreliable target generation~\cite{tang2023optimization, espinoza2023analysis}. At the same time, IK-based controllers remain sensitive to perception noise, visibility constraints, and environmental variation, often producing inefficient or unstable trajectories~\cite{rajendran2024towards}.

To overcome these limitations, we propose a robust end-to-end perception-to-action framework for autonomous greenhouse strawberry harvesting. For perception, we introduce \emph{HRAttnEdge-YOLO26-seg}, a high-resolution, attention-guided, and edge-supervised architecture specifically designed for cluttered fruit scenes. The proposed vision model enhances small-object representation and mask fidelity through three coordinated components: a stride-4 feature branch for preserving fine spatial details, segmentation-path attention for emphasizing fruit-relevant representations, and edge-supervised prototype learning for improving boundary consistency. The resulting segmentation masks are fused with aligned depth information~\cite{11219194}to estimate stable and manipulation-ready 3D fruit targets. For robotic control, we apply deep reinforcement learning (DRL)~\cite{arulkumaran2017deep}, specifically Proximal Policy Optimization (PPO)~\cite{schulman2017proximal} because of its stable optimized characteristics and reliable empirical performance~\cite{kurup2026comparative, elumalai2025proximal}. However, training a physical robot using PPO is often expensive, slow, and non-scalable, requiring extensive iterations~\cite{liu2022digital} and is prone to damage to both hardware and crops~\cite{liu2024review, soleymanzadeh2026towards} owing to its physical trial-and-error process. To avoid these possibility of damage, in this study, we train the  PPO policy algorithm  in the Isaac Lab~\cite{mittal2025isaac} simulation framework for manipulation, which facilitates scalable and domain-randomized learning to generate smooth, target-directed, and continuous joint-position motion for a UR10e robot manipulator without extensive trial-and-error in the real world. This state-of-the-art approach not only reduces development costs but also enables to transfer learned control policy from simulations to physical platforms~\cite{HERNANDEZ2025110161, silveira2025simulation}. Finally, a Robot Operating System (ROS)-based execution framework is developed to integrate perception, 3D target estimation, robotic motion control, and gripper-level harvesting execution into a unified real-world autonomous harvesting pipeline. An overview of the proposed system is illustrated in Figure \ref{fig:overview}.

The main contributions of this study are as follows:
\begin{enumerate}
    \item We propose \emph{HRAttnEdge-YOLO26-seg}, a modified YOLO26-seg~\cite{yolo26_ultralytics} architecture for strawberry instance segmentation in greenhouse scenes, integrating a high-resolution P2 branch, segmentation-path attention, and edge-supervised prototype learning.
    \item A target-conditioned reaching controller is developed for a UR10e manipulator using simulation-based PPO training in Isaac Lab, enabling learning-based motion generation without extensive physical trial-and-error during controller development.
    \item We integrate perception and control into a ROS-based closed-loop harvesting pipeline that converts segmented fruit instances into smoothed 3D robot targets and executes sequential harvesting actions on the physical system.
\end{enumerate}
The proposed framework is evaluated at both the perception and system levels through vision experiments, robot-motion assessment, and real greenhouse harvesting trials.
\section{Methodology}
\label{sec:methodology}
\subsection{Vision Module and 3D Localization}
\label{subsec:perception_module}

\subsubsection{HRAttnEdge-YOLO26-seg: a Modified YOLO26-seg for Strawberry}
\label{subsubsec:modified_yolo26}
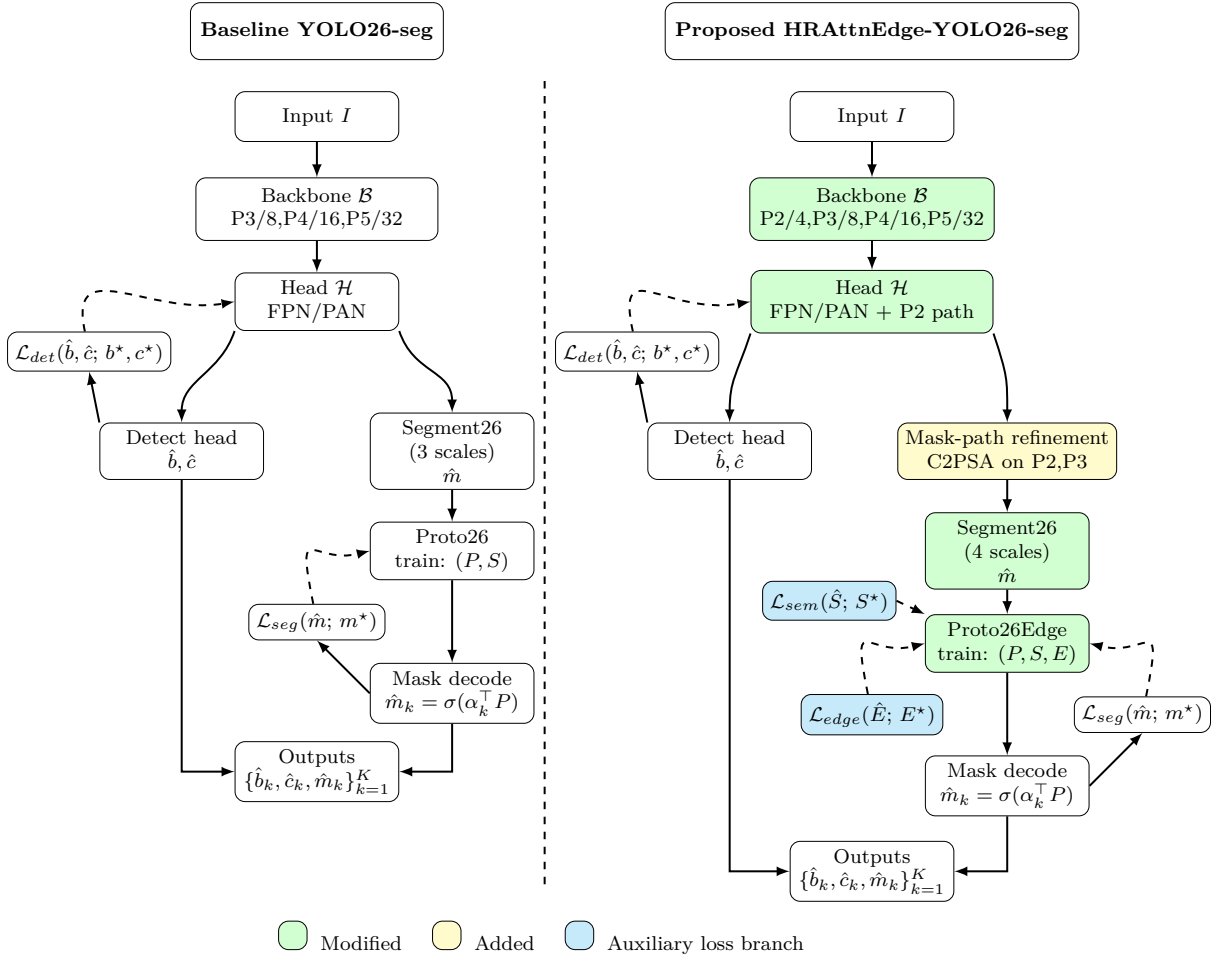
\begin{figure}[t]
\centering
\footnotesize

\resizebox{\linewidth}{!}{%
\begin{tikzpicture}[>=latex,
  block/.style={draw, rounded corners, align=center, inner sep=4pt, minimum width=34mm, minimum height=8mm},
  small/.style={draw, rounded corners, align=center, inner sep=3pt, minimum width=23mm, minimum height=7mm},
  loss/.style={draw, rounded corners, align=center, inner sep=3pt, minimum width=18mm, minimum height=5mm},
  callout/.style={draw, rounded corners, align=left, inner sep=3pt, font=\scriptsize},
  arrow/.style={->, thick},
  dasharrow/.style={->, thick, dashed},
  sepline/.style={dashed, thick},
  added/.style    ={fill=yellow!25},
  modified/.style ={fill=green!18},
  auxiliary/.style  ={fill=cyan!20},
]

\def\xL{0.0}
\def\xR{7.8}
\def\xSep{3.2}
\def\yTop{0.0}

\node[block] (titleL) at (\xL, \yTop) {\textbf{Baseline YOLO26-seg}};
\node[block] (titleR) at (\xR, \yTop) {\textbf{Proposed HRAttnEdge-YOLO26-seg}};

\draw[sepline] (\xSep, \yTop-0.7) -- (\xSep, \yTop-12);

\node[small] (inL)   at (\xL, \yTop-1.2) {Input $I$};
\node[block] (bbL)   at (\xL, \yTop-2.5) {Backbone $\mathcal{B}$\\P3/8,P4/16,P5/32};
\node[small] (headL) at (\xL, \yTop-3.8) {Head $\mathcal{H}$\\FPN/PAN};

\draw[arrow] (inL) -- (bbL);
\draw[arrow] (bbL) -- (headL);

\node[small] (detL) at (\xL-1.9, \yTop-5.9) {Detect head\\$\hat b,\hat c$};
\node[small] (segL) at (\xL+1.9, \yTop-5.9) {Segment26\\(3 scales)\\$\hat m$};

\draw[arrow] (headL.south west) .. controls +(-0.2,-0.6) and +(0,0.6) .. (detL.north);
\draw[arrow] (headL.south east) .. controls +( 0.2,-0.6) and +(0,0.6) .. (segL.north);

\node[small] (protoL)  at (\xL+1.9, \yTop-7.3) {Proto26\\train: $(P,S)$};
\node[small] (decL)    at (\xL+1.9, \yTop-9.3) {Mask decode\\$\hat m_k=\sigma(\alpha_k^\top P)$};
\node[small] (outL)    at (\xL, \yTop-10.4) {Outputs\\$\{\hat b_k,\hat c_k,\hat m_k\}_{k=1}^K$};

\draw[arrow] (segL) -- (protoL);
\draw[arrow] (protoL) -- (decL);

\draw[arrow] (detL.south) |- (outL.west);
\draw[arrow] (decL.south) |- (outL.east);

\node[loss] (lossDetL) at (\xL-3.2, \yTop-4.5) {$\mathcal{L}_{det}(\hat b,\hat c;\,b^\star,c^\star)$};
\node[loss] (lossSegL) at (\xL, \yTop-8.3) {$\mathcal{L}_{seg}(\hat m;\,m^\star)$};

\draw[arrow] (detL.north west) -- (lossDetL.south);
\draw[arrow] (decL.west) -- (lossSegL.south);

\draw[dasharrow] (lossDetL) .. controls +(-0.3,1.4) and +(-1.2,-0.2) .. (headL.west);
\draw[dasharrow] (lossSegL) .. controls +(-0.3,1.5) and +(-.8,-0.2) .. (protoL.west);

\node[small] (inR)   at (\xR, \yTop-1.2) {Input $I$};
\node[block, modified] (bbR)   at (\xR, \yTop-2.5) {Backbone $\mathcal{B}$\\P2/4,P3/8,P4/16,P5/32};
\node[block, modified] (headR) at (\xR, \yTop-3.8) {Head $\mathcal{H}$\\FPN/PAN + P2 path};

\draw[arrow] (inR) -- (bbR);
\draw[arrow] (bbR) -- (headR);

\node[small] (detR) at (\xR-2, \yTop-5.9) {Detect head\\$\hat b,\hat c$};
\node[small, added] (psaR)    at (\xR+1.9, \yTop-5.9) {Mask-path refinement\\C2PSA on P2,P3};

\draw[arrow] (headR.south west) .. controls +(-0.2,-0.6) and +(0,0.6) .. (detR.north);
\draw[arrow] (headR.south east) .. controls +( 0.2,-0.6) and +(0,0.6) .. (psaR.north);

\node[small, modified] (segR) at (\xR+1.9, \yTop-7.3) {Segment26\\(4 scales)\\$\hat m$};
\node[small, modified] (protoR)  at (\xR+1.9, \yTop-8.6) {Proto26Edge\\train: $(P,S,E)$};
\node[small] (decR)    at (\xR+1.9, \yTop-10.6) {Mask decode\\$\hat m_k=\sigma(\alpha_k^\top P)$};
\node[small] (outR)    at (\xR, \yTop-11.8) {Outputs\\$\{\hat b_k,\hat c_k,\hat m_k\}_{k=1}^K$};

\draw[arrow] (psaR) -- (segR);
\draw[arrow] (segR) -- (protoR);
\draw[arrow] (protoR) -- (decR);

\draw[arrow] (detR.south) |- (outR.west);
\draw[arrow] (decR.south) |- (outR.east);

\node[loss] (lossDetR)  at (\xR-3.3, \yTop-4.5) {$\mathcal{L}_{det}(\hat b,\hat c;\,b^\star,c^\star)$};
\node[loss] (lossSegR)  at (\xR+3.8, \yTop-9.6) {$\mathcal{L}_{seg}(\hat m;\,m^\star)$};
\node[loss, auxiliary] (lossEdgeR) at (\xR, \yTop-9.6) {$\mathcal{L}_{edge}(\hat E;\,E^\star)$};
\node[loss, auxiliary] (lossSemR) at (\xR-.6, \yTop-8.0) {$\mathcal{L}_{sem}(\hat S;\,S^\star)$};

\draw[arrow] (detR.north west) -- (lossDetR.south);
\draw[arrow] (decR.east) -- (lossSegR.south);

\draw[dasharrow] (lossDetR) .. controls +(-0.3,1.4) and +(-1.2,-0.2) .. (headR.west);
\draw[dasharrow] (lossSegR) .. controls +( 0.3,1.5) and +( .8,-0.2) .. (protoR.east);
\draw[dasharrow] (lossEdgeR) .. controls +(-0.3,1.5) and +(-.8,-0.2) .. (protoR.west);
\draw[dasharrow] (lossSemR.east) .. controls +(-0.0,.0) and +(-.0,-0.0) .. (protoR.north west);

\node[align=center, inner sep=4pt] (legend) at (\xSep, \yTop-12.75) {%
  \begin{tabular}{@{}c c c@{}}
      \tikz{\node[small, modified, minimum width=4mm, minimum height=4mm] {};}\;  Modified &
      \tikz{\node[small, added, minimum width=4mm, minimum height=4mm] {};}\; Added &
      \tikz{\node[small, auxiliary , minimum width=4mm, minimum height=4mm] {};}\; Auxiliary loss branch
  \end{tabular}
};
\end{tikzpicture}%
} 

\vspace{-2mm}
\caption{Side-by-side comparison of baseline YOLO26-seg and the proposed HRAttnEdge-YOLO26-seg. The proposed model integrates a stride-4 segmentation feature (+P2), applies C2PSA to the P2 and P3 mask-path features, replaces Proto26 with Proto26Edge, and introduces auxiliary semantic and boundary supervision during training. Solid arrows denote forward inference flow, whereas dashed arrows indicate gradient propagation from the corresponding loss terms during backpropagation.}
\label{fig:baseline_vs_proposedYOLOsegn}
\vspace{-3mm}
\end{figure}

We address strawberry instance segmentation in cluttered and occluded greenhouse scenes. Given an RGB image $I \in \mathbb{R}^{H\times W \times 3}$, the model predicts $K$ instances, ${(\hat{b}_k, \hat{c}_k, \hat{m}_k)}_{k=1}^K$, where $\hat{b}_k \in \mathbb{R}^4$ is the bounding box, $\hat{c}_k \in {1,\dots,C}$ is the class label, and $\hat{m}_k \in {0,1}^{H\times W}$ is the binary mask.
Our baseline is YOLO26-seg~\cite{yolo26_ultralytics}, a lightweight one-stage instance segmentation model whose head uses three pyramid levels: P3/8, P4/16, and P5/32. While efficient, this design is less effective in agriculture scenes, where masks are degraded due to small visible fruit regions, leaf and stem clutter, partial occlusion, and fine boundaries around the calyx and fruit edge. To address this, we propose HRAttnEdge-YOLO26-seg: a high-resolution attention-guided and edge-supervised YOLO26-seg, a modification of YOLO26-seg that improves high-resolution mask quality. The final design includes: (i) P2/4 integration into the segmentation head to preserve stride-4 spatial detail; (ii) Cross Stage Partial with Spatial Attention (C2PSA) applied to the segmentation path of P2/4 and P3/8 features to strengthen boundary-aware representation with limited overhead; and (iii) a multi-scale fused prototype generator with boundary supervision (Proto26Edge), which adds an edge head and auxiliary edge loss to the prototype branch. The model is optimized for both detection and segmentation, with particular emphasis on high-IoU mask quality, where boundary precision is most critical. Figure \ref{fig:baseline_vs_proposedYOLOsegn} highlights the differences between the baseline and the proposed model.

\paragraph{ P2/4 high-resolution branch.}
The baseline head uses $\{P_3,P_4,P_5\}$ features at strides $\{8,16,32\}$ for segmentation. Strawberries often contain fine contours (calyx, specular edges, thin occluded boundaries) that might poorly represented at stride 8. We therefore introduce an additional stride-4 feature level $P_2/4$ and feed it into the segmentation head, yielding the four-scale feature set $\{P_2,P_3,P_4,P_5\}$.
Let $\mathbf{F}_l$ denote the refined feature map at pyramid level $l$.
The newly introduced $P_2$ feature is formed by upsampling the refined $P_3$ feature and concatenating it with the corresponding backbone feature:
\begin{equation}
\mathbf{F}_{2} = \phi_{2}\!\left(\left[\operatorname{Up}(\mathbf{F}_{3}), \mathbf{B}_{2}\right]\right),
\label{eq:p2_fusion}
\end{equation}
where $\mathbf{B}_{2}$ is the backbone feature at P2, $\operatorname{Up}(\cdot)$ denotes upsampling, $[\cdot,\cdot]$ denotes channel concatenation, and $\phi_{2}(\cdot)$ denotes the convolutional refinement block. This additional branch retains fine spatial details that would otherwise be lost in lower-resolution features.

\paragraph{Segmentation-path targeted attention.}
To sharpen mask representation without increasing computation across the entire network, we apply attention only along the mask-critical path. Specifically, we insert C2PSA on the stride-4 and stride-8 head outputs. Let $\mathcal{A}(\cdot)$ denote the C2PSA attention operator. The refined attended features are:
\begin{equation}
\widetilde{\mathbf{F}}_{2} = \mathcal{A}(\mathbf{F}_{2}),
\qquad
\widetilde{\mathbf{F}}_{3} = \mathcal{A}(\mathbf{F}_{3}),
\label{eq:attention_features}
\end{equation}
where $\widetilde{\mathbf{F}}_{2}$ and $\widetilde{\mathbf{F}}_{3}$ are used in subsequent segmentation refinement and mask-coefficient prediction. This choice targets the two resolutions that most strongly affect boundary precision: $\mathbf{F}_{2}$ provides pixel-level geometry, while $\mathbf{F}_{3}$ offers a detail semantics balance.
We also implemented a guard that prevents zero-head configurations by enforcing  a valid head count in PSA:
\begin{equation}
n_h = \max\!\left(1, \left\lfloor \frac{d}{64} \right\rfloor \right),
\label{eq:psa_heads}
\end{equation}
where $d$ denotes the PSA internal channel width. 

\paragraph{Multi-scale prototype learning with edge supervision.}
The standard prototype branch, Proto26, has been replaced by a multi-scale fused prototype module, denoted as \emph{Proto26Edge}.
Given the feature set $\{\widetilde{\mathbf{F}}_2,\widetilde{\mathbf{F}}_3,\mathbf{F}_4,\mathbf{F}_5\}$, multi-scale prototype fusion is expressed as
\begin{equation}
\mathbf{F}_{proto} = \psi\!\left(
\widetilde{\mathbf{F}}_2 + 
\sum_{l=3}^{5} \operatorname{Up}_{l \rightarrow 2}\!\left(g_l(\mathbf{F}_l)\right)
\right),
\label{eq:proto_fusion}
\end{equation}
where $g_l(\cdot)$ is a $1\times1$ channel-alignment operator, $\operatorname{Up}_{l \rightarrow 2}(\cdot)$ upsamples each feature to the $P_2$ spatial resolution, and $\psi(\cdot)$ denotes the final fusion convolution.
From $\mathbf{F}_{proto}$, the model predicts the shared mask prototypes:
\begin{equation}
\mathbf{P} = h_{proto}(\mathbf{F}_{proto}) \in \mathbb{R}^{N_p \times H_p \times W_p},
\label{eq:prototypes}
\end{equation}
where $N_p$ is the number of prototype channels.
Each instance mask is reconstructed by linearly combining the prototypes with an instance-specific mask coefficient vector $\mathbf{c}_i$:
\begin{equation}
\mathbf{M}_i = \sigma\!\left(\sum_{k=1}^{N_p} c_{i,k}\,\mathbf{P}_k\right),
\label{eq:mask_reconstruction}
\end{equation}
where $\sigma(\cdot)$ denotes the sigmoid activation and $\mathbf{P}_k$ is the $k$-th prototype channel.

In addition to prototype learning, the fused feature $\mathbf{F}_{proto}$ is used to predict an auxiliary boundary map:
\begin{equation}
\mathbf{E}_{pred} = h_{edge}(\mathbf{F}_{proto}),
\label{eq:edge_pred}
\end{equation}
where $h_{edge}(\cdot)$ is a lightweight edge head.
This auxiliary branch explicitly encourages crisp object boundaries during training.

\paragraph{Loss formulation.}
\label{para:seg_loss}

The proposed model is optimized using a composite objective consisting of detection, instance segmentation, auxiliary semantic supervision, and auxiliary boundary supervision. The total training loss is defined as
\begin{equation}
\mathcal{L}_{\text{vision}} =
\mathcal{L}_{\text{box}} +
\mathcal{L}_{\text{cls}} +
\mathcal{L}_{\text{dfl}} +
\lambda_{\text{seg}} \mathcal{L}_{\text{seg}} +
\lambda_{\text{sem}} \mathcal{L}_{\text{sem}} +
\lambda_{\text{edge}} \mathcal{L}_{\text{edge}},
\label{eq:vision_total_loss}
\end{equation}
where $\mathcal{L}_{\text{box}}$ denotes the bounding-box regression loss, $\mathcal{L}_{\text{cls}}$ the classification loss, and $\mathcal{L}_{\text{dfl}}$ the distribution focal loss inherited from the YOLO26 detection framework. The term $\mathcal{L}_{\text{seg}}$ is the instance mask loss, $\mathcal{L}_{\text{sem}}$ is the auxiliary semantic segmentation loss produced by the prototype branch, and $\mathcal{L}_{\text{edge}}$ is an auxiliary boundary loss used to encourage sharper mask contours.

For boundary supervision, a binary foreground map $\mathbf{G}$ is first formed by taking the union of all ground-truth instance masks in each image. The corresponding edge target is then generated by a morphological gradient operation,
\begin{equation}
\mathbf{E}_{gt} = \operatorname{Dilate}(\mathbf{G}) - \operatorname{Erode}(\mathbf{G}),
\label{eq:edge_target}
\end{equation}
which highlights contour pixels around the foreground boundary. Let $\hat{\mathbf{E}}$ denote the predicted edge logit map from the auxiliary edge branch. The edge loss is computed as
\begin{equation}
\mathcal{L}_{\text{edge}} =
\lambda_{\mathrm{bce}}\,\mathcal{L}_{\mathrm{BCE}}(\hat{\mathbf{E}}, \mathbf{E}_{gt})
+
\lambda_{\mathrm{dice}}\,\mathcal{L}_{\mathrm{Dice}}(\hat{\mathbf{E}}, \mathbf{E}_{gt}),
\label{eq:edge_loss}
\end{equation}
where $\mathcal{L}_{\mathrm{BCE}}$ and $\mathcal{L}_{\mathrm{Dice}}$ denote the binary cross-entropy and Dice losses, respectively. This auxiliary edge branch is used only during training to improve boundary awareness and is not required for standard instance-mask inference.

\subsubsection{3D fruit-center localization}
\label{subsubsec:3d_localization}

The perception module was implemented as a ROS node that subscribes to RGB images, corresponding depth frames, and camera intrinsics.
For each segmented strawberry instance from the HRAttnEdge-YOLO26-seg model, the 2D center point $(u,v)$ was estimated from the mask polygon coordinates.
The aligned depth value at the center pixel was then used to recover the 3D fruit-center position in the camera coordinate frame.

Given pixel coordinates $(u,v)$, depth value $Z_c$, and camera intrinsics $(f_x,f_y,c_x,c_y)$, the 3D point $(X_c,Y_c,Z_c)$ in the camera frame is computed by pinhole back-projection:
\begin{equation}
X_c = \frac{(u-c_x) Z_c}{f_x}, \qquad
Y_c = \frac{(v-c_y) Z_c}{f_y}, \qquad
Z_c = Z_c.
\label{eq:backprojection}
\end{equation}

This process generates a 3D target for each segmented fruit. Only valid depth values were used; detections with $\leq 75\% $ Intersection over Union (IoU) and invalid depth were discarded.

\subsubsection{Temporal association and smoothing}
\label{subsubsec:tracking_smoothing}

3D fruit targets in greenhouse scenes may fluctuate across frames because of depth noise, small segmentation variations, and partial occlusions. To reduce target jitter before robot execution, a lightweight temporal association and smoothing scheme was applied.

Let $\mathbf{p}_t^{(j)}$ denote the 2D center of segmented mask $j$ in frame $t$, and let $\hat{\mathbf{p}}_{t-1}^{(k)}$ denote the most recent 2D center stored in active track $k$. For each new detection, the nearest existing track is found in the image plane as
\begin{equation}
k^\ast(j)=
\arg\min_{k\in\mathcal{K}_{t-1}}
\left\|
\mathbf{p}_t^{(j)}-\hat{\mathbf{p}}_{t-1}^{(k)}
\right\|_2,
\label{eq:nn_match}
\end{equation}
where $\mathcal{K}_{t-1}$ is the set of active tracks from the previous frame. The assignment is accepted only if the corresponding pixel distance is less than the threshold $\tau_p$;  otherwise, a new track is initialized.

After association, the 3D fruit center, obtained from \eqref{eq:backprojection} for mask $j$, is appended to the buffer of its matched track $k^\ast(j)$. For each track,  a buffer of recent 3D estimates is maintained, and once the buffer is filled, the smoothed target is computed by temporal averaging:
\begin{equation}
\bar{\mathbf{q}}^{(k)}
\frac{1}{N_b}
\sum_{n=1}^{N_b}
\mathbf{q}_n^{(k)},
\label{eq:smoothing}
\end{equation}
where $\mathbf{q}_n^{(k)}$ is the $n$-th buffered 3D estimate associated with track $k$, and $N_b$ is the buffer length ($N_b=15$ in our experiment). The averaged point $\bar{\mathbf{q}}^{(k)}$ is then published as the final smoothed target for downstream robot control.

\subsection{DRL-based Target Reaching Policy in Simulation}
\label{subsec:drl_module}

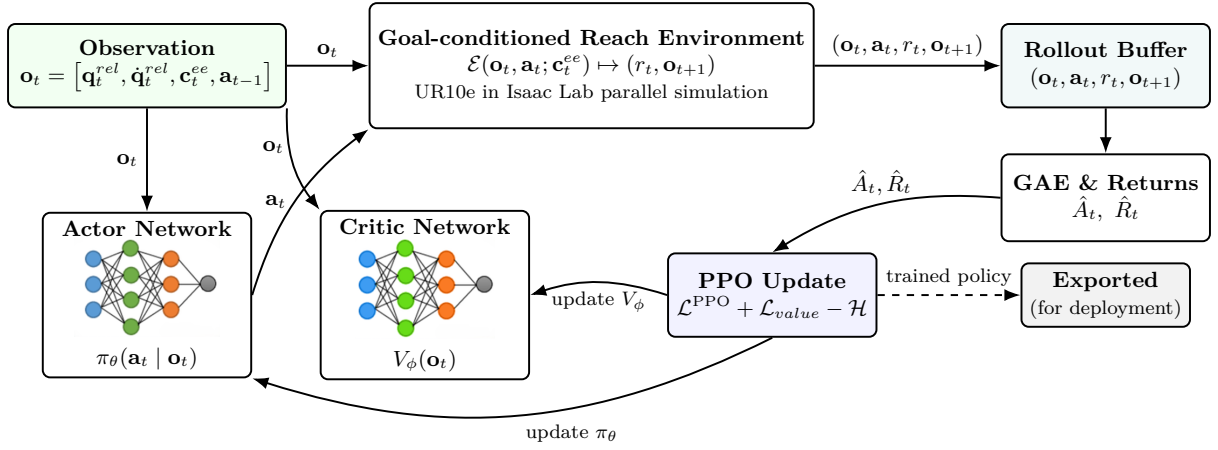
\begin{figure*}[t]
\centering
\resizebox{\textwidth}{!}{%
\begin{tikzpicture}[
    font=\small,
    >=Latex,
    box/.style={
        draw,
        rounded corners=3pt,
        thick,
        align=center,
        minimum height=1.45cm,
        minimum width=3.4cm
    },
    smallbox/.style={
        draw,
        rounded corners=3pt,
        thick,
        align=center,
        minimum height=1.25cm,
        minimum width=3.0cm
    },
    line/.style={->, thick},
    dashedline/.style={->, thick, dashed}
]

\node[box, minimum width=4.0cm, minimum height=1.2cm, fill=green!5] (obs) at (0,2.8)
{\textbf{Observation}\\
$\mathbf{o}_t=
\left[
\mathbf{q}_t^{rel},
\dot{\mathbf{q}}_t^{rel},
\mathbf{c}_t^{ee},
\mathbf{a}_{t-1}
\right]$};

\node[box, minimum width=4.8cm, minimum height=1.8cm] (env) at (6.4,2.8)
{\textbf{Goal-conditioned Reach Environment}\\
$\mathcal{E}\!\left(\mathbf{o}_t,\mathbf{a}_t;\mathbf{c}_t^{ee}\right)
\mapsto
\left(r_t,\mathbf{o}_{t+1}\right)$\\
\footnotesize UR10e in Isaac Lab parallel simulation};

\node[box, minimum width=3cm, minimum height=1.2cm, fill=teal!5] (rollout) at (13.8,2.8)
{\textbf{Rollout Buffer}\\
$\left(\mathbf{o}_t,\mathbf{a}_t,r_t,\mathbf{o}_{t+1}\right)$};

\node[smallbox] (gae) at (13.8,0.9) {\textbf{GAE \& Returns}\\
$\hat{A}_t,\ \hat{R}_t$};

\node[box, minimum width=3cm, minimum height=1.2cm, fill=blue!5] (ppo) at (9,-0.5)
{\textbf{PPO Update}\\
$\mathcal{L}^{\text{PPO}}+\mathcal{L}_{value}-\mathcal{H}$};

\node[box, minimum width=3.0cm, minimum height=1.7cm] (actor) at (0,-0.5)
{\textbf{Actor Network}\\
\includegraphics[width=2.1cm]{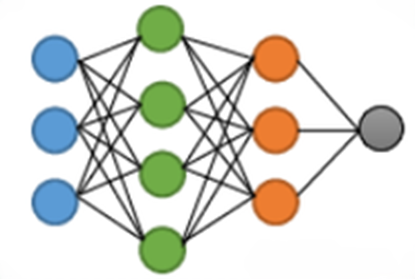}\\
$\pi_{\theta}(\mathbf{a}_t\mid\mathbf{o}_t)$};

\node[box, minimum width=3.0cm, minimum height=1.7cm] (critic) at (4.0,-0.5)
{\textbf{Critic Network}\\
\includegraphics[width=2cm]{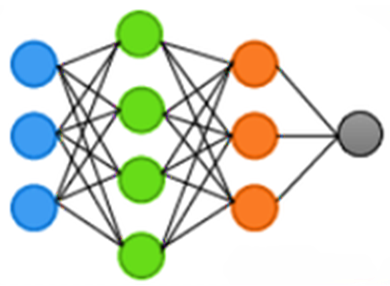}\\
$V_{\phi}(\mathbf{o}_t)$};

\node[box, minimum width=2cm, minimum height=0.8cm, fill=gray!10] (deploy) at (13.8,-0.5)
{\textbf{Exported}\\
\footnotesize (for deployment)};

\draw[line] (obs.east) -- node[above] {$\mathbf{o}_t$} (env.west);
\draw[line] (obs.south) -- node[left] {$\mathbf{o}_t$} (actor.north);
\draw[line] (obs.south east) to[bend right=20] node[midway, above left] {$\mathbf{o}_t$} (critic.north west);

\draw[line] (actor.east) to[bend left=18] node[midway, left] {$\mathbf{a}_t$} (env.south west);

\draw[line] (env.east) -- node[above] {$\left(\mathbf{o}_t,\mathbf{a}_t,r_t,\mathbf{o}_{t+1}\right)$} (rollout.west);
\draw[line] (rollout.south) -- (gae.north);
\draw[line] (gae.west) to[bend right=18] node[midway, above]{$\hat{A}_t,\hat{R}_t$} (ppo.north);

\draw[line] (ppo.west) to[bend right=18] node[midway, below] {\footnotesize update $V_{\phi}$} (critic.east);
\draw[line] (ppo.south) to[bend left=22] node[midway, below right] {\footnotesize update $\pi_{\theta}$} (actor.south east);

\draw[dashedline] (ppo.east) -- node[above] {\footnotesize trained policy} (deploy.west);

\end{tikzpicture}%
}
\caption{PPO-based reaching policy learning for the UR10e manipulator in Isaac Lab. At each time step, the policy receives the observation vector $\mathbf{o}_t=[\mathbf{q}_t^{rel},\dot{\mathbf{q}}_t^{rel},\mathbf{c}_t^{ee},\mathbf{a}_{t-1}]$, where $\mathbf{q}_t^{rel}$ and $\dot{\mathbf{q}}_t^{rel}$ denote relative joint positions and velocities, $\mathbf{c}_t^{ee}$ is the commanded end-effector pose, and $\mathbf{a}_{t-1}$ is the previous action. The goal-conditioned reach environment is denoted by $\mathcal{E}(\mathbf{o}_t,\mathbf{a}_t;\mathbf{c}_t^{ee})$, indicating that the current observation and action are propagated under the commanded end-effector pose to produce the immediate reward $r_t$ and the next observation $\mathbf{o}_{t+1}$. The actor network outputs the continuous joint-position action $\mathbf{a}_t$, while the critic estimates the value function $V_{\phi}(\mathbf{o}_t)$. Interaction tuples $\left(\mathbf{o}_t,\mathbf{a}_t,r_t,\mathbf{o}_{t+1}\right)$ are stored in the rollout buffer and used to compute the generalized advantage estimate (GAE) $\hat{A}_t$ and return $\hat{R}_t$. PPO then updates the policy and value networks. At each check-point the trained actor is exported as a TorchScript policy for ROS-based deployment.}
\label{fig:ppo_architecture}
\end{figure*}

The target reaching controller was trained in Isaac Lab environment built around the UR10e robotic arm and a sampled end-effector pose command. At each episode, a target pose was generated within a predefined reachable workspace, and the control objective was to drive the end-effector from its initial configuration to the commanded pose within a fixed episode horizon. Policy learning utilized Isaac Lab's standard reinforcement learning workflow with PPO, in which parallel simulation environments were used. In this experiment, 64 environments were employed in parallel to gather rollout trajectories for policy and value updates. Figure~\ref{fig:ppo_architecture} summarizes the overall PPO-based learning pipeline used for UR10e reaching policy training.

\subsubsection{Observation and action spaces}
\label{subsubsec:obs_action}

The policy input was defined as a compact observation vector composed of robot proprioception, task command, and action history:
\begin{equation}
\mathbf{o}_t=
\left[
\mathbf{q}_t^{rel},
\ \dot{\mathbf{q}}_t^{rel},
\ \mathbf{c}_t^{ee},
\ \mathbf{a}_{t-1}
\right],
\label{eq:observation_vector_revised}
\end{equation}
where $\mathbf{q}_t^{rel}\in\mathbb{R}^{6}$ and $\dot{\mathbf{q}}_t^{rel}\in\mathbb{R}^{6}$ denote the relative joint positions and joint velocities of the UR10e, respectively, $\mathbf{c}_t^{ee}$ denotes the commanded end-effector pose, and $\mathbf{a}_{t-1}$ is the previous action applied by the policy. In this formulation, the pose command serves as the task-conditioning term that specifies the desired end-effector goal at time step $t$.

The policy output was a continuous joint-position action
\begin{equation}
\mathbf{a}_t \in \mathbb{R}^{6},
\label{eq:action_vector_revised}
\end{equation}
corresponding to the six revolute joints of the UR10e arm. These actions were passed through Isaac Lab's joint-position action interface after scaling, allowing the policy to generate incremental motions toward the commanded pose. The inclusion of the previous action in Eq.~\eqref{eq:observation_vector_revised} encouraged temporal consistency and reduced abrupt changes in joint commands.

\subsubsection{Reward formulation}
\label{subsubsec:reward_formulation_revised}

The reward function combined target-reaching objectives with motion regularization. Let $\mathbf{x}^{ee}_t\in\mathbb{R}^{3}$ and $\mathbf{x}^{cmd}_t\in\mathbb{R}^{3}$ denote the current and commanded end-effector positions, respectively. The position tracking error was defined as
\begin{equation}
d_t = \left\|\mathbf{x}^{ee}_t-\mathbf{x}^{cmd}_t\right\|_2.
\label{eq:pos_error_revised}
\end{equation}

To encourage both coarse convergence and accurate final positioning, the reward included two complementary position terms. The first term penalized Euclidean distance to the target:
\begin{equation}
r_{pos}(t) = -\,d_t,
\label{eq:reward_pos_revised}
\end{equation}
which encouraged global movement toward the commanded position. The second term provided fine-grained shaping near the goal through a bounded exponential form:
\begin{equation}
r_{fine}(t) = \exp\!\left(-\frac{d_t}{\sigma}\right),
\label{eq:reward_fine_revised}
\end{equation}
where $\sigma>0$ controlled the sharpness of reward concentration around the target. This term increased the reward sensitivity in the near-goal region, thereby improving terminal precision.

In addition to positional alignment, the controller was trained to track the commanded end-effector orientation. Let $\mathbf{R}^{ee}_t$ and $\mathbf{R}^{cmd}_t$ denote the current and commanded end-effector rotations, respectively, and let $\Delta\theta_t$ denote the corresponding orientation error. The orientation reward was written abstractly as
\begin{equation}
r_{ori}(t) = -\,\Delta\theta_t,
\label{eq:reward_ori_revised}
\end{equation}
where $\Delta\theta_t$ was computed from the mismatch between the current and target end-effector orientations.

To regularize motion, two penalties were introduced. The first penalized rapid action changes:
\begin{equation}
r_{act}(t) = -\left\|\mathbf{a}_t-\mathbf{a}_{t-1}\right\|_2^2,
\label{eq:reward_action_rate_revised}
\end{equation}
and the second penalized high joint velocities:
\begin{equation}
r_{vel}(t) = -\left\|\dot{\mathbf{q}}_t\right\|_2^2.
\label{eq:reward_joint_vel_revised}
\end{equation}

The total reward at each time step was therefore defined as
\begin{equation}
r_t
=
w_1 r_{pos}(t)
+
w_2 r_{fine}(t)
+
w_3 r_{ori}(t)
+
w_4 r_{act}(t)
+
w_5 r_{vel}(t),
\label{eq:total_reward_revised}
\end{equation}
where $w_1,\dots,w_5$ are scalar weighting coefficients specified in the environment configuration. This reward formulation guided the policy toward accurate goal-reaching while discouraging oscillatory and dynamically aggressive motions.

\subsubsection{PPO optimization}
\label{subsubsec:ppo_revised}

The reaching policy was optimized using PPO with the RSL-RL~\cite{schwarke2025rslrl} backend. Let $\pi_\theta(\mathbf{a}_t\mid\mathbf{o}_t)$ denote the stochastic policy parameterized by $\theta$, and let $V_\phi(\mathbf{o}_t)$ denote the value function parameterized by $\phi$. PPO updates the policy by maximizing the clipped surrogate objective:
\begin{equation}
\mathcal{L}^{\text{PPO}}(\theta)=
\mathbb{E}_t
\left[
\min
\left(
\rho_t(\theta)\,\hat{A}_t,\;
\operatorname{clip}\!\left(\rho_t(\theta),1-\epsilon,1+\epsilon\right)\hat{A}_t
\right)
\right],
\label{eq:ppo_objective_revised}
\end{equation}
where
\begin{equation}
\rho_t(\theta)=
\frac{\pi_\theta(\mathbf{a}_t\mid\mathbf{o}_t)}
{\pi_{\theta_{\mathrm{old}}}(\mathbf{a}_t\mid\mathbf{o}_t)}
\label{eq:importance_ratio_revised}
\end{equation}
is the importance sampling ratio between the updated policy and the behavior policy used to collect rollout samples, $\hat{A}_t$ is the estimated advantage, and $\epsilon$ is the clipping parameter.

The advantage term was computed using generalized advantage estimation (GAE):
\begin{equation}
\hat{A}_t
=
\sum_{l=0}^{T-t-1}
(\gamma\lambda)^l
\delta_{t+l},
\label{eq:gae_revised}
\end{equation}
where $\gamma$ is the discount factor, $\lambda$ is the GAE decay parameter, and the temporal-difference residual is
\begin{equation}
\delta_t
=
r_t
+
\gamma V_\phi(\mathbf{o}_{t+1})
-
V_\phi(\mathbf{o}_t).
\label{eq:td_error_revised}
\end{equation}

In addition to the clipped policy objective, PPO included a value regression term and an entropy regularization term. The overall optimization objective can be written as
\begin{equation}
\mathcal{L}_{total}^{\text{PPO}}
=
-\mathcal{L}^{\text{PPO}}
+
c_v\,\mathcal{L}_{value}
-
c_e\,\mathcal{H}\!\left[\pi_\theta(\cdot\mid\mathbf{o}_t)\right],
\label{eq:ppo_total_loss_revised}
\end{equation}
where $c_v$ and $c_e$ are scalar coefficients controlling the contributions of the value loss and entropy bonus, respectively. The entropy term encouraged sufficient exploration during training, while the value term stabilized learning by improving return estimation.

In this study, the core PPO algorithm itself was not modified. Instead, the contribution of the control module lies in training a simulation-based reaching policy for the UR10e arm within an Isaac Lab environment, formulating the goal-conditioned reward function, and then integrating the exported policy into the full robotic strawberry harvesting pipeline for real-system execution.

\begin{figure*}[t]
\centering
\resizebox{\linewidth}{!}{%
\begin{tikzpicture}[
    x=.5cm,y=.5cm,
    >=Latex,
    font=\footnotesize,
    node distance=0.9cm and 0.9cm,
    block/.style={
        draw,
        rounded corners=3pt,
        align=center,
        minimum width=1.5cm,
        minimum height=.8cm,
        thick
    },
    arrow/.style={->, thick},
    note/.style={align=center, font=\footnotesize}
]

\node[block, fill=blue!5] (cam) {Camera node\\(RGB-D)};
\node[block, fill=green!5, right=.6cm of cam] (vision) {Vision node\\(HRAttnEdge-YOLO26-seg\\ +\\smooth 3D target)};
\node[block, fill=cyan!5, right=.6cm of vision] (harvest) {Harvest controller\\(task sequencing)};
\node[block, fill=green!5, right=.6cm of harvest] (policy) {Policy controller\\(PPO)};
\node[block, fill=teal!5 , right=.6cm of policy] (robot) {Command interface\\(reach streamer)};
\node[block, fill=blue!5, below=.6cm of harvest] (gripper) {Gripper controller\\(SG-b-h)};

\draw[arrow] (cam) -- (vision);
\draw[arrow] (vision) -- (harvest);
\draw[arrow] (harvest) -- (policy);
\draw[arrow] (policy) -- (robot);

\draw[arrow] (harvest) -- (gripper);
\draw[arrow] (gripper.east) -| ([yshift=-0.15cm]robot.south);

\begin{pgfonlayer}{background}
\node[
    draw,
    dashed,
    thick,
    fill=gray!10,
    rounded corners=4pt,
    inner xsep=0.3cm,
    inner ysep=0.3cm,
    fit=(cam)(vision)(harvest)(policy)(robot)(gripper)
] (rosbox) {};
\end{pgfonlayer}

\node[
    anchor=south west,
    font=\footnotesize\bfseries,
    inner sep=1.5pt
] at ([xshift=0.15cm,yshift=0.15cm]rosbox.south west) {ROS~1};

\end{tikzpicture}
}
\caption{ROS-based integration architecture for greenhouse strawberry harvesting. The vision node publishes smoothed 3D fruit targets, the harvesting controller manages task sequencing, the policy controller generates target-conditioned robot actions, and the gripper branch executes deterministic grasp and release commands.}
\label{fig:ros_integration}
\end{figure*}
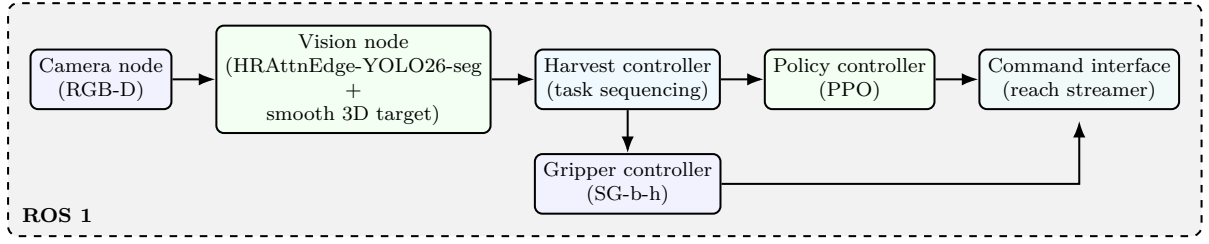

\subsection{ROS-based Integration for Harvesting}
\label{subsec:ros_integration}

For physical deployment, we developed a ROS-based framework that integrates perception, policy-based reaching, and high-level harvesting execution within a single runtime pipeline (Figure \ref{fig:ros_integration}). The system consists of three functional layers: (i) a perception layer for segmentation and 3D fruit-center estimation, (ii) a policy-control layer that loads the exported reaching policy and computes robot actions, and (iii) a task-execution layer that coordinates the harvesting sequence. This modular design separates perception, control, and operations logic, enabling closed-loop execution while simplifying debugging and module-level validation.

The perception node publishes temporally smoothed strawberry centers as ROS \texttt{PointStamped} messages. These targets are mapped from the camera to the UR10e operating frame using a calibrated coordinate transformation. Let $(X_c, Y_c, Z_c)$ denote the fruit center in the camera frame. The corresponding robot-frame target $(X_r, Y_r, Z_r)$ is computed as:
\begin{equation}
\begin{bmatrix}
X_r\\Y_r\\Z_r
\end{bmatrix}
=
\mathbf{T}
\begin{bmatrix}
X_c\\Y_c\\Z_c\\1
\end{bmatrix},
\label{eq:camera_to_robot}
\end{equation}
where $\mathbf{T}$ denotes the effective rigid transformation and offset mapping from perception space to robot workspace.

The exported PPO policy generates target-conditioned joint-position actions for the UR10e arm. Because the learned policy produces spatial joint commands without explicit temporal scheduling, a command-streaming layer was introduced between policy inference and robot execution. Let $\Delta \mathbf{q}_t$ denote the policy-generated joint displacement at step $t$. The minimum feasible execution time is chosen as
\begin{equation}
\Delta t_t
=
\max_i
\left(
\frac{|\Delta q_{t,i}|}{\dot{q}_{i,\max}}
\right),
\label{eq:time_parameterization}
\end{equation}
where $\Delta q_{t,i}$ is the commanded displacement of joint $i$ and $\dot{q}_{i,\max}$ is its maximum allowable velocity. The pair $(\Delta \mathbf{q}_t,\Delta t_t)$ is then encoded into a ROS \texttt{JointTrajectory} message for execution by the UR10e controller, allowing the learned target-directed motion to be applied safely on hardware.

A halt mechanism was further incorporated to suppress steady-state oscillation near the target. During continuous execution, small residual policy corrections can produce high-frequency motion updates even after the end-effector has effectively converged. Rather than using a simple position-threshold stop, which may terminate motion prematurely, the streamer monitors command transitions online and engages a halt state when the resulting motion demand exceeds a predefined safety threshold near convergence. This reduces end-effector vibration and stabilizes the final reaching state during physical harvesting.

Each harvesting trial starts from a predefined home position, where the perception node localizes ripe strawberries in the workspace. A sequential harvesting routine then prioritizes the nearest detected berry, executes the learned reaching policy, and performs a downward pull-based detachment. After transferring the fruit to the collection basket, the robot proceeds to the next identified instance in the current frame. This process continues until no valid targets remain, at which point the robot returns to the home position and re-scans the scene if needed.

Gripper actuation was handled separately from the reaching policy through deterministic commands during grasping and release. This separation assigns target-reaching to the learned controller while preserving a stable and repeatable grasp-and-release routine at the end effector.

\section{Experiment}
\label{experiment}
\subsection{Experimental Setup}
\label{subsec:experimental_platform}
The vision module was trained on a Linux workstation running Ubuntu 22.04 LTS equipped with dual Intel Xeon Gold processors (3.0~GHz, 24 cores, 35.75~MB cache), 256~GB RAM, and an NVIDIA RTX A6000 GPU with 48~GB VRAM. All experiments were conducted in the Ultralytics training framework using mixed-precision (AMP) to improve throughput. Unless otherwise stated, training used an input resolution of $640\times640$, batch size 16, and a maximum of 200 epochs with early stopping (patience = 15). Optimization used \textit{AdamW} with an initial learning rate of $2\times10^{-4}$ under a cosine learning-rate schedule. Standard photometric and geometric augmentations were enabled (HSV jitter, rotation/translation/scale/shear, and horizontal flips), together with composition-based augmentation suitable for instance segmentation (mosaic, MixUp, and segment copy–paste); mosaic was disabled during the final 10 epochs to stabilize late-stage boundary refinement. For the proposed \texttt{Proto26Edge} branch, auxiliary boundary supervision was applied throughout training with an edge-loss gain of 0.2. To ensure a fair comparison between baseline and modified architectures, all models were trained from scratch without pretrained weights.

The control policy was trained on the same workstation using Isaac Lab's manager-based reinforcement learning workflow with the Robotics Student League -- Reinforcement Learning (RSL-RL) backend on top of Isaac Sim 5.1.0. Isaac Lab provided the vectorized environment wrappers required for large-scale parallel policy training, along with policy export utilities.

The experimental deployment platform consisted of a RB-VOGUI mobile robot (Robotnik Automation, Paterna, Spain) with an onboard Intel i7-based computer running Linux, a UR10e robotic arm (Universal Robot, Odense, Denmark) mounted on top of the mobile platform, an SG-b-h soft gripper (OnRobot, Odense, Denmark) attached to the tool center point, and an Intel RealSense 435i RGB-D camera mounted near the end-effector for close-range perception. During execution, the perception, policy inference, and task-level harvesting controllers were operated under ROS~1 on the onboard computer, while the UR10e arm was interfaced through the PolyScope 5.19 control environment.  The experiment took place in a strawberry greenhouse at the Texas A\&M AgriLife Research Center in Dallas, TX, USA.

\begin{table*}[t]
\small
\centering
\setlength{\tabcolsep}{5pt}
\renewcommand{\arraystretch}{1.15}
\caption{Summary of the datasets used in this study.}
\label{tab:datasets}
\begin{tabular}{@{}lccccccl@{}}
\toprule
\multirow{2}{*}{Dataset} 
& \multirow{2}{*}{\#Images} 
& \multirow{2}{*}{\shortstack[c]{\#Instances\\(approx.)}} 
& \multicolumn{3}{c}{Split ratio (\%)} 
& \multicolumn{1}{c}{Split size (\# images)} 
& \multirow{2}{*}{Label} \\
\cmidrule(lr){4-6} \cmidrule(lr){7-7}
& & & Train & Val & Test & Train / Val / Test & \\
\midrule
Self-collected ($D_{\mathrm{self}}$) 
& 647 
& 1494 
& 70.0 
& 15.0 
& 15.0 
& 453 / 97 / 97 
& ripe\_strawberry \\

Public ($D_{\mathrm{pub}}$)~\cite{PEREZBORRERO2020105736} 
& 3100 
& 197,997 
& 90.3 
& 3.2 
& 6.5 
& 2800 / 100 / 200 
& strawberry \\
\bottomrule
\end{tabular}
\end{table*}

\subsection{Datasets for Vision Model}
\label{sub:datasets}
Table \ref{tab:datasets} presents a summary of the datasets utilized for training and evaluation of the vision model in this study. We evaluated our proposed HRAttnEdge-YOLO26-seg model using two distinct strawberry datasets. The first dataset consisted of self-collected images of greenhouse-grown strawberries, captured with an Intel RealSense 435i RGB-D camera (Intel Corporation, Santa Clara, CA, USA). This dataset contained a total of 653 color images,  with each image containing approximately 2.3 instances of ripe strawberries. Annotations were created using CVAT~\cite{Sekachev2020}, following the YOLO segmentation format, which employs normalized polygon masks. This dataset was divided into 70\% for training, 15\% for validation, and 15\% for testing.
    
The second dataset we used is StrawDI (The Strawberry Digital Images Data Set)~\cite{PEREZBORRERO2020105736}, which is publicly available. This dataset contains 3,100 color images, with an average of approximately 6.4 instances of strawberries per image. Each image has a corresponding ground truth grayscale polygon mask,  where the background is coded as 0, and strawberries are labeled with values ranging from 1 to n (where n is the total number of instances). We converted these grayscale labels into the YOLO segmentation format. The dataset publisher divided the images as follows: 2,800 images (approximately 90.32\%) for training, 100 (3.23\%) for validation, and 200 (6.45\%) for testing. We preserved this same split to train, validate, and evaluate our model, as well as to conduct comparisons with the state-of-the-art.

In both datasets, the validation set was utilized to guide model convergence during training. Ablation studies and comparisons with state-of-the-art methods were conducted using the held-out test set completely separately.

\subsection{Evaluation Metrics}
\label{subsec:eval_metrics}
\subsubsection{Performance Assessment of Vision}
The performance of the vision models was evaluated using standard metrics. Detection and segmentation performance were assessed on instance bounding boxes and masks, respectively. We report precision ($P$) and recall ($R$), where overlap between prediction and ground truth is measured by intersection over union (IoU). We also report $\mathrm{mAP}_{50}$ and $\mathrm{mAP}_{75}$ respectively, together with $\mathrm{mAP}_{50:95}$, which averages AP over IoU thresholds from 0.50 to 0.95 in steps of 0.05~\cite{lin2014microsoft, everingham2010pascal}.

\subsubsection{Motion and Harvesting Assessment}
\label{subsec:motion_assessment}
We evaluated the integrated robotic harvesting pipeline from two complementary perspectives: (i) motion quality of the robot during target-reaching segments, and (ii) task-level success at harvesting trials of the target, while the localized target comes from the vision module. The motion analysis characterizes efficiency and smoothness of the executed trajectories, whereas the task-level metrics reflect whether the system can reliably reach and harvest strawberries under closed-loop operation.

\paragraph{Trajectory logging and smoothing.}
To assess motion quality, the transformation between the robot base frame and tool frame was sampled periodically using Transform Library-2 (TF2) during each movement segment. This produced a sequence of Cartesian samples-
\begin{equation}
\mathcal{T} = \left\{(\tau_i, \mathbf{p}_i)\right\}_{i=1}^{N},
\qquad
\mathbf{p}_i = [x_i, y_i, z_i]^\top,
\label{eq:trajectory_samples}
\end{equation}
where $\tau_i$ denotes the sampling time and $\mathbf{p}_i$ denotes the tool position.

To reduce the effects of transform noise and irregular timestamps, the trajectory was interpolated onto a uniform time grid and smoothed using a moving-average filter. For path-length estimation, the smoothed trajectory was further simplified using the Ramer--Douglas--Peucker~\cite{douglas1973algorithms, ramer1972iterative} algorithm. The resulting trajectory is denoted as-
\begin{equation}
\widetilde{\mathbf{p}}(t) = [\widetilde{x}(t), \widetilde{y}(t), \widetilde{z}(t)]^\top.
\label{eq:smoothed_traj}
\end{equation}

\paragraph{Motion metrics.}
Each motion segment was characterized using the metrics listed in Table~\ref{tab:motion_metrics}. Let $t_{\text{start}}$ and $t_{\text{end}}$ denote the start and end times of a segment. The segment duration is
\begin{equation}
T=t_{\text{end}}-t_{\text{start}}.
\label{eq:duration}
\end{equation}

The straight-line distance between the initial and final tool positions is
\begin{equation}
D_{\text{straight}}=
\left\|
\widetilde{\mathbf{p}}(t_{\text{end}})
-
\widetilde{\mathbf{p}}(t_{\text{start}})
\right\|_2.
\label{eq:straight_distance}
\end{equation}

If the simplified trajectory is represented by $M$ ordered points $\{\mathbf{s}_m\}_{m=1}^{M}$, the executed trajectory length is
\begin{equation}
L_{\text{traj}}=
\sum_{m=1}^{M-1}
\left\|
\mathbf{s}_{m+1}-\mathbf{s}_m
\right\|_2.
\label{eq:traj_length}
\end{equation}

The mean speed is then computed as
\begin{equation}
\bar{v}=\frac{L_{\text{traj}}}{T}.
\label{eq:mean_speed}
\end{equation}

To quantify motion smoothness, velocity, acceleration, and jerk were obtained by numerical differentiation of the smoothed trajectory:
\begin{equation}
\mathbf{v}(t)=\frac{d\widetilde{\mathbf{p}}(t)}{dt},
\qquad
\mathbf{a}(t)=\frac{d\mathbf{v}(t)}{dt},
\qquad
\mathbf{j}(t)=\frac{d\mathbf{a}(t)}{dt}.
\label{eq:jerk_def}
\end{equation}
The jerk magnitude is
\begin{equation}
j(t)=\|\mathbf{j}(t)\|_2.
\label{eq:jerk_mag}
\end{equation}

The root-mean-square jerk is defined as
\begin{equation}
J_{\text{rms}}=
\sqrt{
\frac{1}{T}
\int_{t_{\text{start}}}^{t_{\text{end}}}
j(t)^2\,dt
},
\label{eq:rms_jerk}
\end{equation}
which summarizes the overall smoothness of the motion over the full segment. To reduce sensitivity to isolated numerical spikes, the peak jerk was reported using a robust percentile statistic,
\begin{equation}
J_{\text{peak}}^{\text{robust}}=
Q_q\!\left(j(t)\right),
\label{eq:robust_peak_jerk}
\end{equation}
where $Q_q(\cdot)$ denotes the $q$-th percentile of the jerk magnitude sequence.

\paragraph{Task-level success metrics.}
In addition to motion quality, we evaluated whether the robot could successfully complete the reaching and harvesting tasks (Table \ref{tab:motion_metrics}). Let $N_r$ denote the total number of reaching attempts. For the $k$-th attempt, define the binary indicator
\begin{equation}
s_k^{\text{reach}}=
\begin{cases}
1, & \text{if } d_k^{\text{final}} \le \varepsilon_r \text{ and the target is reached before timeout},\\
0, & \text{otherwise},
\end{cases}
\label{eq:reach_indicator}
\end{equation}
where $d_k^{\text{final}}$ is the final end-effector-to-target position error and $\varepsilon_r$ is the reaching tolerance ($\varepsilon_r =2$  cm in this experiment). The reaching success rate is then
\begin{equation}
R_{\text{success}}=
\frac{1}{N_r}
\sum_{k=1}^{N_r}
s_k^{\text{reach}}
\times 100\%.
\label{eq:reach_success}
\end{equation}

Similarly, let $N_h$ denote the total number of harvesting attempts. For the $k$-th target, define
\begin{equation}
s_k^{\text{harvest}}=
\begin{cases}
1, & \text{if the fruit is successfully grasped, detached, and deposited},\\
0, & \text{otherwise}.
\end{cases}
\label{eq:harvest_indicator}
\end{equation}
The harvesting success rate is computed as
\begin{equation}
H_{\text{success}}=
\frac{1}{N_h}
\sum_{k=1}^{N_h}
s_k^{\text{harvest}}
\times 100\%.
\label{eq:harvest_success}
\end{equation}

\begin{table}[t]
\centering
\small
\caption{Motion and harvesting metrics used for system-level evaluation.}
\label{tab:motion_metrics}
\begin{tabular}{p{3.3cm} p{2.0cm} p{7.2cm}}
\toprule
\textbf{Metric} & \textbf{Symbol} & \textbf{Definition} \\
\midrule
Duration & $T$ & Execution time of a motion segment \\
Straight-line distance & $D_{\text{straight}}$ & Euclidean distance between segment start and end positions \\
Trajectory length & $L_{\text{traj}}$ & Cumulative length of the executed path \\
Mean speed & $\bar{v}$ & Average path speed over the segment \\
RMS jerk & $J_{\text{rms}}$ & Overall smoothness of motion \\
Robust peak jerk & $J_{\text{peak}}^{\text{robust}}$ & High-percentile jerk magnitude \\
Reaching success rate & $R_{\text{success}}$ & Percentage of attempts that reached the commanded target within tolerance \\
Harvesting success rate & $H_{\text{success}}$ & Percentage of attempts that completed grasping, detachment, and deposition \\
\bottomrule
\end{tabular}
\end{table}

\subsection{Results}
\label{sub:results}
\begin{figure*}[t]
\centering
\includegraphics[width=.96\textwidth]{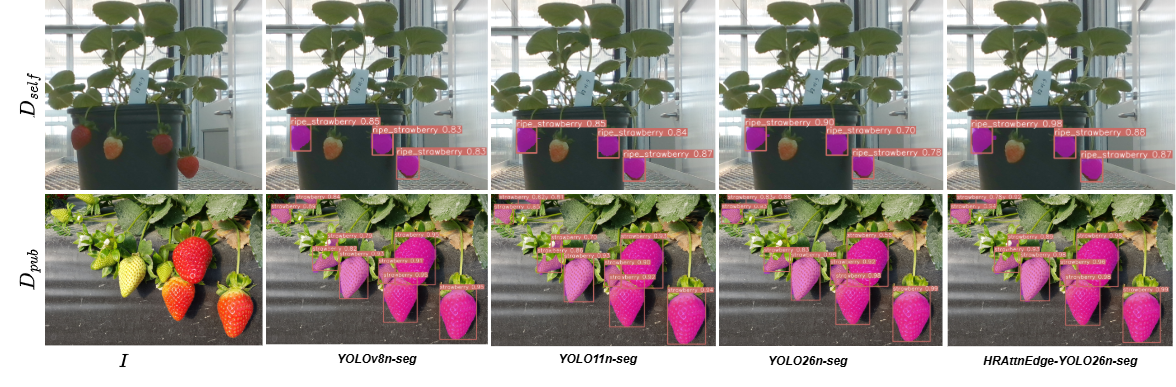}
\caption{Qualitative comparison of strawberry instance segmentation results at the nano ($n$) scale on the self-collected (top, D\textsubscript{self}) and public (bottom, D\textsubscript{pub}) datasets. From left to right, each row shows the input image ($I$), YOLOv8n-seg, YOLO11n-seg, YOLO26n-seg, and the proposed HRAttnEdge-YOLO26-seg. The proposed model produces more complete fruit masks and tighter boundary delineation, even for small, partially occluded, and cluttered scenes.}
\label{fig:vision_compare}
\end{figure*}

\subsubsection{Vision System Results}
\paragraph{Comparative Experiments.}
Table~\ref{tab:main_results} compares the proposed HRAttnEdge-YOLO26-seg with the baseline YOLO26-seg and prior YOLO-seg generations across nano ($n$), small ($s$), and medium ($m$) scales on both the self-collected (D\textsubscript{self}) and public (D\textsubscript{pub}) datasets. Overall, the proposed model showed the strongest and most consistent performance across detection and segmentation metrics. This behavior is important for robotic harvesting because improvements in $\mathrm{mAP}_{75}$ and $\mathrm{mAP}_{50:95}$ reflect tighter instance delineation and more reliable fruit boundary localization, which directly support stable 3D target generation for downstream reaching and harvesting.

In our self-collected dataset, the proposed model demonstrated the best overall performance across all three scales. The $m$ variant yielded the strongest results, achieving detection and segmentation scores of 99.1 and 98.9, respectively, at $\mathrm{mAP}_{50}$. Compared to the baseline YOLO26-seg, our model improved segmentation performance by 14.4\%, 13.6\%, and 9.5\% on the $n$, $s$, and $m$ scales, respectively, at $\mathrm{mAP}_{75}$. Additionally, it exhibited the highest precision and recall. This trend was consistent across all scales at $\mathrm{mAP}_{50:95}$, indicating that the improvements were not limited to basic foreground recovery but also extended to more precise mask formation, even under stricter IoU conditions.

The public dataset posed greater challenges for all models, evident in the generally lower scores and the more significant drop under stricter overlap criteria. Yet, the proposed model remained the strongest overall performer. In comparison to the baseline YOLO26-seg, mask $\mathrm{mAP}_{50}$ increased from 73.8 to 84.6 (14.6\%$\uparrow$), 79.3 to 89.2 (12.5\%$\uparrow$), and 81.9 to 92.1 (12.4\%$\uparrow$) at the $n$, $s$, and $m$ scales, respectively. Similar improvements were observed for $\mathrm{mAP}_{75}$ and $\mathrm{mAP}_{50:95}$. Furthermore, the method demonstrated higher precision and recall, indicating enhanced cross-dataset robustness, which is essential for model generalizability. In our settings, the baseline YOLO26-seg consistently underperformed compared to YOLOv8-seg and YOLO11-seg on the public dataset. However, the proposed modifications rendered the YOLO26 branch not only competitive but also superior overall. Figure~\ref{fig:vision_compare} presents qualitative comparisons of strawberry instance segmentation results produced by different models at the nano ($n$) scale.

\begin{table*}[t]
\small
\centering
\setlength{\tabcolsep}{4pt}
\renewcommand{\arraystretch}{1.15}
\caption{Comparison of model performance on the self-collected and public datasets.}
\label{tab:main_results}
\begin{tabular}{lllcccccccc}
\toprule
\multirow{3}{*}{Model} 
& \multirow{3}{*}{Scale} 
& \multirow{3}{*}{Dataset} 
& \multirow{3}{*}{P} 
& \multirow{3}{*}{R} 
& \multicolumn{3}{c}{Bbox (detection)} 
& \multicolumn{3}{c}{Mask (segmentation)} \\

\cmidrule(lr){6-8}\cmidrule(lr){9-11}
& & & & 
& mAP\textsubscript{50} 
& mAP\textsubscript{75} 
& mAP\textsubscript{50:95} 
& mAP\textsubscript{50} 
& mAP\textsubscript{75} 
& mAP\textsubscript{50:95} \\

\midrule

\multirow{6}{*}{YOLOv8-seg~\cite{yolov8_ultralytics}}
& \multirow{2}{*}{n} & D\textsubscript{self} & 91.2 & 92.5  & 91.4 & 83.0 & 67.8 & 90.6 & 74.0 & 62.8 \\
& & D\textsubscript{pub}  & 84.5 & 73.4 & 81.8 & 74.2 & 69.9 & 81.3 & 72.0 & 51.4 \\
& \multirow{2}{*}{s} & D\textsubscript{self} & 92.5 & 93.1  & 92.1 & 85.7 & 68.9 & 92.1 & 79.8 & 66.9 \\
& & D\textsubscript{pub}  & 85.4 & 74.6 & 90.0 & 81.6 & 62.9 & 89.6 & 78.5 & 56.3 \\
& \multirow{2}{*}{m} & D\textsubscript{self} & 94.7 & 94.0 & 94.3 & 86.7 & 80.2 & 92.3 & 79.2 & 66.6 \\
& & D\textsubscript{pub}  & 86.7 & 86.6 & 85.4 & 72.7 & 68.8 & 85.4 & 72.3 & 62.8 \\
\midrule

\multirow{6}{*}{YOLO11-seg~\cite{yolo11_ultralytics}}
& \multirow{2}{*}{n} & D\textsubscript{self} & 91.8 & 92.6  & 92.4 & 83.8 & 71.5 & 91.9 & 81.4 & 63.6 \\
& & D\textsubscript{pub}  & 84.3 & 73.8 & 82.7 & 72.7 & 69.3 & 82.3 & 74.2 & 52.9 \\
& \multirow{2}{*}{s} & D\textsubscript{self} & 93.9 & 92.8  & 91.2 & 87.9 & 73.1 & 92.6 & 83.2 & 67.3 \\
& & D\textsubscript{pub}  & 84.8 & 74.5 & 89.5 & 72.5 & 61.3 & 82.7 & 68.0 & 54.1 \\
& \multirow{2}{*}{m} & D\textsubscript{self} & 95.7 & 91.9 & 94.1 & 88.5 & 79.5 & 93.7 & 77.2 & 68.8 \\
& & D\textsubscript{pub}  & 90.9 & 86.3 & 88.8 & 75.5 & 62.0 & 84.1 & 74.9 & 60.6 \\
\midrule

\multirow{6}{*}{YOLO26-seg~\cite{yolo26_ultralytics}}
& \multirow{2}{*}{n} & D\textsubscript{self} & 89.9 & 83.5 & 89.8 & 79.3 & 64.1 & 89.2 & 72.3 & 63.8 \\
& & D\textsubscript{pub}  & 68.0 & 71.4 & 73.8 & 70.2 & 57.6 & 73.8 & 61.9 & 50.9 \\
& \multirow{2}{*}{s} & D\textsubscript{self} & 90.4 & 91.3 & 92.4 & 74.7 & 65.2 & 91.7 & 73.2 & 64.7 \\
& & D\textsubscript{pub}  & 77.9 & 78.5 & 79.3 & 68.4 & 57.5 & 76.4 & 67.5 & 55.1 \\
& \multirow{2}{*}{m} & D\textsubscript{self} & 93.0 & 91.6 & 96.6 & 78.8 & 67.7 & 95.6 & 76.7 & 66.0 \\
& & D\textsubscript{pub}  & 84.7 & 80.4 & 81.9 & 73.5 & 62.0 & 79.1 & 69.2 & 61.6 \\
\midrule


\rowcolor{lightgrayrow}
& n & D\textsubscript{self} & 96.5 & 94.9 & 96.0 & 85.8 & 73.1 & 95.4 & 82.7 & 71.0 \\
\rowcolor{lightgrayrow}
&   & D\textsubscript{pub}  & 89.0 & 85.4 & 84.6 & 73.2 & 61.6 & 84.4 & 72.0 & 60.9 \\
\rowcolor{lightgrayrow}
& s & D\textsubscript{self} & 97.3 & 95.9 & 97.7 & 94.6 & 76.1 & 97.7 & 83.2 & 74.2 \\
\rowcolor{lightgrayrow}
&   & D\textsubscript{pub}  & 89.3 & 86.2 & 89.2 & 72.2 & 62.1 & 89.4 & 73.5 & 62.5 \\
\rowcolor{lightgrayrow}
& m & D\textsubscript{self} & 98.5 & 98.6 & 99.1 & 98.3 & 85.3 & 98.9 & 84.0 & 76.3 \\
\rowcolor{lightgrayrow}
\multirow{-6}{*}{\cellcolor{lightgrayrow}\shortstack[l]{HRAttnEdge-YOLO26-seg\\(ours)}}
&   & D\textsubscript{pub}  & 91.8 & 87.6 & 92.1 & 83.1 & 65.3 & 92.0 & 80.3 & 65.1 \\
\bottomrule
\end{tabular}
\end{table*}

\paragraph{Ablation Experiments.}
Table~\ref{tab:ablation_results} summarizes the effect of progressively adding the proposed components to YOLO26-seg. The performance gain follows a clear architectural trend. Adding the P2 branch yields modest but consistent improvement by preserving finer spatial detail for small fruit regions and thin boundaries. Adding segmentation-path attention provides further gain by refining these high-resolution mask features more effectively. The largest improvement appears after introducing EdgeProto, indicating that the final gain comes from the combined effect of fine-resolution features, targeted attention, and boundary-aware prototype learning rather than from scale alone. The trade-off is increased overhead. The rise in GFLOPs for the +P2 and +P2+Attn. variants are expected, since extending the segmentation head from P3--P5 to P2--P5 shifts more computation to the high-resolution stride-4 feature map, and the added attention further increases processing along this dense mask path. After introducing EdgeProto, however, GFLOPs decrease relative to +P2+Attn. because the final prototype branch uses a more efficient multi-scale fusion design, even though inference time remains higher due to branch complexity and feature-fusion overhead. Relative to the baseline, the full model increases size from 6.4 to 6.6 MB (3.1\%$\uparrow$), 23.0 to 23.5 MB (2.2\%$\uparrow$), and 54.0 to 56.1 MB (3.9\%$\uparrow$) for the $n$, $s$, and $m$ scales, respectively, with corresponding increases in parameter count and inference time. Even so, the accuracy gain is substantial, particularly under stricter mask criteria, and the model remains practical for deployment. This is crucial for edge-oriented robotic execution, as lightweight models are often preferred. In this context, the nano and small variants are appealing for resource-constrained platforms, while the medium variant provides the highest accuracy when computational resources are less limited. In our integrated robotic system, we utilized the nano variant of the proposed HRAttnEdge-YOLO26-seg vision model for harvesting operations.

\begin{table*}[t]
\small
\centering
\caption{Ablation study of HRAttnEdge-YOLO26-seg on self-collected (D\textsubscript{self}) and public (D\textsubscript{pub}) datasets across nano ($n$), small ($s$), and medium ($m$) scales. Components are progressively added starting from the YOLO26-seg baseline: the high-resolution P2 branch, segmentation-path attention, and EdgeProto.}
\label{tab:ablation_results}
\setlength{\tabcolsep}{4.5pt}
\begin{tabular}{@{}lcccccccc@{}}
\toprule
\multirow{2}{*}{Model variant}&\multirow{2}{*}{Scale}&\multirow{2}{*}{Dataset}&\multicolumn{2}{c}{Segmentation performance}&\multirow{2}{*}{\begin{tabular}[c]{@{}c@{}}Size\\(MB)\end{tabular}}&\multirow{2}{*}{\begin{tabular}[c]{@{}c@{}}\#Params\\(M)\end{tabular}}&\multirow{2}{*}{\begin{tabular}[c]{@{}c@{}}Inference\\(ms)\end{tabular}}&\multirow{2}{*}{\begin{tabular}[c]{@{}c@{}}GFLOPs\\@640px\end{tabular}}\\
\cmidrule(lr){4-5}
&&&mAP\textsubscript{50}&mAP\textsubscript{75}&&&&\\
\midrule
\multirow{6}{*}{\begin{tabular}[c]{@{}l@{}}Baseline\\(YOLO26-seg~\cite{yolo26_ultralytics})\end{tabular}}&\multirow{2}{*}{n}& D\textsubscript{self} &89.2&72.3&\multirow{2}{*}{6.4}&\multirow{2}{*}{2.69}&\multirow{2}{*}{3.7}&\multirow{2}{*}{8.9}\\
&&D\textsubscript{pub}&73.8&61.9&&&&\\
&\multirow{2}{*}{s}&D\textsubscript{self}&91.4&73.2&\multirow{2}{*}{23.0}&\multirow{2}{*}{10.36}&\multirow{2}{*}{4.7}&\multirow{2}{*}{29.7}\\
&&D\textsubscript{pub}&76.4&67.5&&&&\\
&\multirow{2}{*}{m}&D\textsubscript{self}&95.6&76.7&\multirow{2}{*}{54.0}&\multirow{2}{*}{23.51}&\multirow{2}{*}{5.8}&\multirow{2}{*}{121.5}\\
&&D\textsubscript{pub}&79.1&69.2&&&&\\
\midrule
\multirow{6}{*}{+P2}&\multirow{2}{*}{n}&D\textsubscript{self}&89.8&76.7&\multirow{2}{*}{6.5}&\multirow{2}{*}{2.91}&\multirow{2}{*}{12.6}&\multirow{2}{*}{23.1}\\
&&D\textsubscript{pub}&74.4&62.3&&&&\\
&\multirow{2}{*}{s}&D\textsubscript{self}&91.7&73.9&\multirow{2}{*}{23.1}&\multirow{2}{*}{11.16}&\multirow{2}{*}{17.21}&\multirow{2}{*}{86.4}\\
&&D\textsubscript{pub}&78.6&69.3&&&&\\
&\multirow{2}{*}{m}&D\textsubscript{self}&96.2&77.6&\multirow{2}{*}{54.5}&\multirow{2}{*}{26.67}&\multirow{2}{*}{23.3}&\multirow{2}{*}{329.0}\\
&&D\textsubscript{pub}&85.0&75.8&&&&\\
\midrule
\multirow{6}{*}{+P2+Attn.}&\multirow{2}{*}{n}&D\textsubscript{self}&92.3&76.8&\multirow{2}{*}{6.5}&\multirow{2}{*}{2.93}&\multirow{2}{*}{8.8}&\multirow{2}{*}{24.1}\\
&&D\textsubscript{pub}&75.4&63.3&&&&\\
&\multirow{2}{*}{s}&D\textsubscript{self}&94.0&75.4&\multirow{2}{*}{23.2}&\multirow{2}{*}{11.28}&\multirow{2}{*}{10.4}&\multirow{2}{*}{90.4}\\
&&D\textsubscript{pub}&78.8&70.1&&&&\\
&\multirow{2}{*}{m}&D\textsubscript{self}&96.4&78.7&\multirow{2}{*}{55.1}&\multirow{2}{*}{27.17}&\multirow{2}{*}{23.7}&\multirow{2}{*}{344.8}\\
&&D\textsubscript{pub}&87.1&77.3&&&&\\
\midrule
\multirow{6}{*}{\begin{tabular}[c]{@{}l@{}}+P2+Attn.+EdgeProto\\(ours)\end{tabular}}&\multirow{2}{*}{n}&D\textsubscript{self}&95.4&82.7&\multirow{2}{*}{6.6}&\multirow{2}{*}{2.97}&\multirow{2}{*}{14.9}&\multirow{2}{*}{14.6}\\
&&D\textsubscript{pub}&84.4&72.0&&&&\\
&\multirow{2}{*}{s}&D\textsubscript{self}&97.7&83.2&\multirow{2}{*}{23.5}&\multirow{2}{*}{11.43}&\multirow{2}{*}{16.7}&\multirow{2}{*}{53.8}\\
&&D\textsubscript{pub}&89.4&73.5&&&&\\
&\multirow{2}{*}{m}&D\textsubscript{self}&98.9&84.0&\multirow{2}{*}{56.1}&\multirow{2}{*}{27.73}&\multirow{2}{*}{33.1}&\multirow{2}{*}{198.2}\\
&&D\textsubscript{pub}&92.0&80.3&&&&\\
\bottomrule
\end{tabular}
\end{table*}

\begin{figure*}[t]
\centering
\includegraphics[width=.82\textwidth]{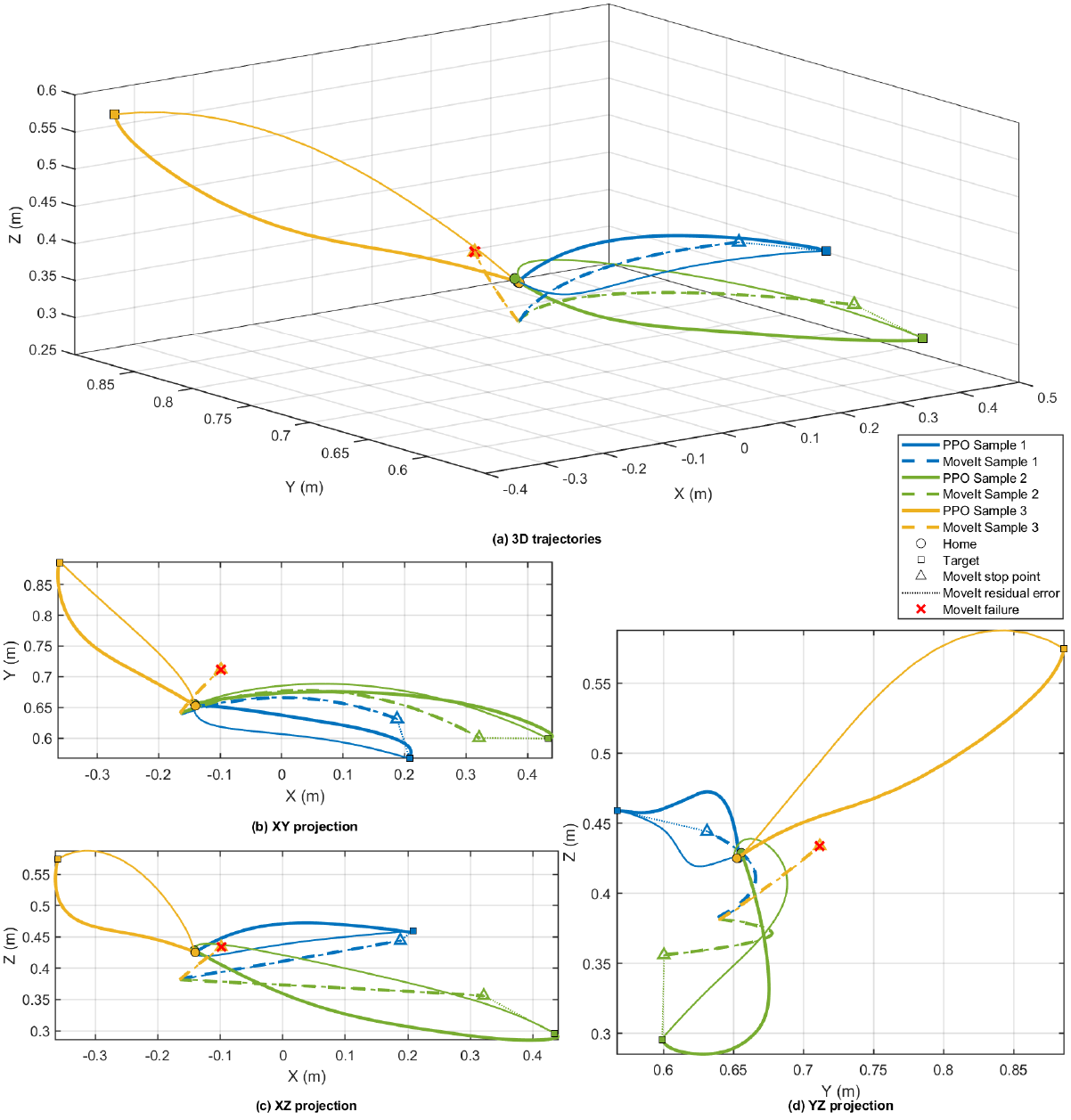}
\caption{Representative in-house trajectories for qualitative comparison between the PPO-based controller and IK-based MoveIt: (a) 3D trajectories, (b) XY projection, (c) XZ projection, and (d) YZ projection. Each sample shows an actual Home$\rightarrow$Target approach and the corresponding Target$\rightarrow$Home return motion, where the target denotes the selected strawberry. Solid curves denote PPO and dashed curves denote IK-based MoveIt. Circular markers denote home, squares mark targets, hollow triangles show actual MoveIt stops, and dotted segments represent residual error. The red cross marks a representative MoveIt failure case.}
\label{fig:inhouse_traj_compare}
\end{figure*}

\subsubsection{Motion and Harvesting Results}
\label{subsec:motion_results}
\paragraph{In-house motion comparison between PPO and IK-based control.}
Before greenhouse deployment, we compared our PPO-based reaching policy controller with an IK-based MoveIt controller using 120 artificial strawberry targets in an in-house setting. This comparison was intended to assess controller behavior under controlled conditions before conducting harvest trials in the more constrained greenhouse workspace. The evaluation focused on the two shared motion primitives, namely Home$\rightarrow$Strawberry and Strawberry$\rightarrow$Basket. As shown in Table~\ref{tab:inhouse_motion_compare}, the PPO controller produced substantially more stable motion than IK across both primitives. For Home$\rightarrow$Strawberry motion, PPO showed lower mean duration and far smaller variability than IK, together with markedly lower RMS jerk. A similar trend was observed for Strawberry$\rightarrow$Basket transfer, where PPO again maintained lower jerk and much tighter dispersion. Although IK-based control showed slightly higher mean speed, this did not correspond to better motion quality, because it was accompanied by substantially larger variation in execution time, path length, and smoothness. These quantitative differences are consistent with the representative trajectories shown in Figure ~\ref{fig:inhouse_traj_compare}, where PPO follows smoother and more coherent approach-and-return paths, while IK-based MoveIt exhibits more irregular motion, larger residual target-reaching error, and a representative failure case. In practice, this behavior was also consistent with repeated full-arm rotations, mid-motion stops, failure to reach the target, and occasional contact with the robot base observed during testing. These in-house results indicate that PPO generated more predictable and dynamically smoother motion, whereas IK behaved less reliably and was therefore not considered suitable for greenhouse harvesting.

\begin{table*}[t]
\centering
\small
\setlength{\tabcolsep}{4.5pt}
\renewcommand{\arraystretch}{1.12}
\caption{In-house motion performance comparison between the PPO controller and IK-based MoveIt using 120 artificial strawberry targets. Values are reported as mean $\pm$ standard deviation. Here, H $\rightarrow$ S denotes Home to target Strawberry, and S $\rightarrow$ B denotes Strawberry to Basket.}
\label{tab:inhouse_motion_compare}
\begin{tabular}{llccccc}
\toprule
\textbf{Controller} & \textbf{Primitive} &
\makecell{\textbf{Duration}\\\textbf{(s)}} &
\makecell{\textbf{Trajectory}\\\textbf{length (m)}} &
\makecell{\textbf{Mean}\\\textbf{speed (m/s)}} &
\makecell{\textbf{RMS}\\\textbf{jerk (m/s$^3$)}} &
\makecell{\textbf{Peak}\\\textbf{jerk (m/s$^3$)}} \\
\midrule
\multirow{2}{*}{PPO}
& H $\rightarrow$ S & $5.95 \pm 1.63$ & $1.69 \pm 0.40$ & $0.29 \pm 0.07$ & $693.8 \pm 157.7$ & $5730.8 \pm 782.4$ \\
& S $\rightarrow$ B & $3.76 \pm 1.83$ & $1.55 \pm 0.41$ & $0.24 \pm 0.07$ & $685.3 \pm 323.7$ & $5393.4 \pm 867.7$ \\
\midrule
\multirow{2}{*}{IK-based}
& H $\rightarrow$ S & $6.76 \pm 6.89$ & $0.94 \pm 1.04$ & $0.33 \pm 0.21$ & $1000.9 \pm 417.2$ & $5374.8 \pm 1338.4$ \\
& S $\rightarrow$ B & $3.88 \pm 6.58$ & $0.82 \pm 0.63$ & $0.26 \pm 0.22$ & $1040.7 \pm 465.2$ & $5579.2 \pm 1043.5$ \\
\bottomrule
\end{tabular}
\end{table*}

\begin{figure*}[t]
\centering
\includegraphics[width=.50\textwidth]{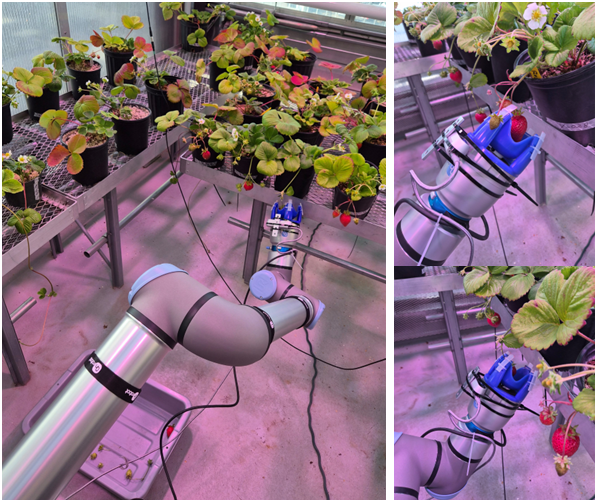}
\caption{Demonstration of greenhouse strawberry harvesting using integrated vision and trajectory control. For harvesting operations, the nano-scale HRAttnEdge-YOLO26-seg and PPO-based reaching policy were utilized.}
\label{fig:harvest_demo}
\end{figure*}

\begin{figure*}[t]
\centering
\includegraphics[width=.82\textwidth]{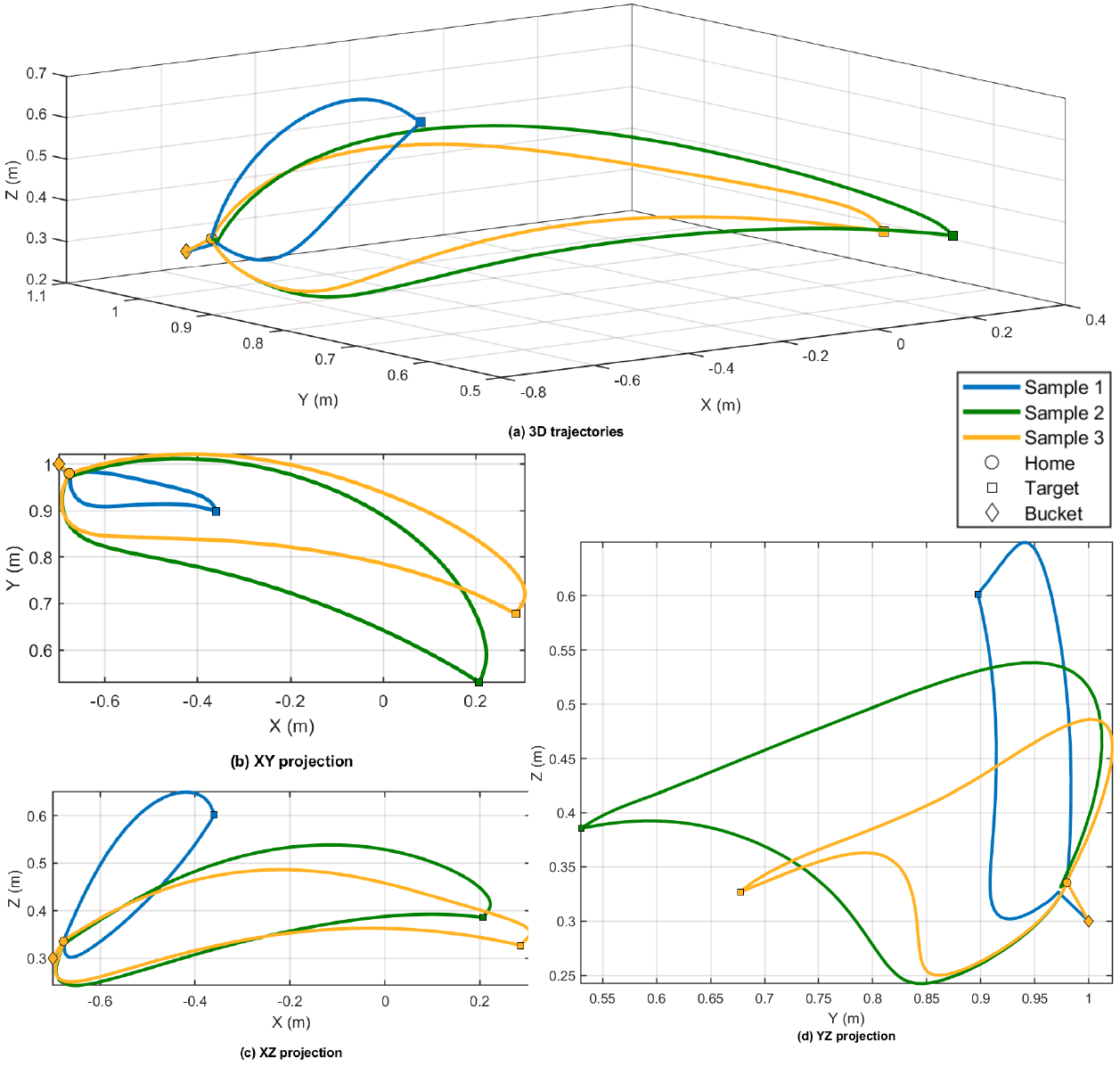}
\caption{Representative greenhouse harvesting trajectories of the PPO-based controller: (a) 3D trajectories, (b) XY projection, (c) XZ projection, and (d) YZ projection. Each sample shows an actual Home$\rightarrow$Target$\rightarrow$Bucket motion.}
\label{fig:greenhouse_ppo_traj}
\end{figure*}

\paragraph{Motion and harvesting performance of the PPO controller in greenhouse settings.}
The greenhouse harvesting experiments were conducted using only the PPO-based controller, as IK-based method was not sufficiently reliable for operation in the constrained greenhouse workspace. A total of 281 strawberries were harvested over 130 trials initiated from the home position, while the vision system enabled a multi-target harvesting strategy with a mean of 2.16 localized berries per trial in this experiment. As summarized in Table~\ref{tab:greenhouse_motion_ppo}, the Home$\rightarrow$Strawberry motion was the most stable primitive, showing the lowest mean speed and the lowest jerk variation, which indicates controlled target approach in the cluttered greenhouse workspace. The Pull motion was shorter and faster, but also more dynamically aggressive, as reflected by its substantially higher RMS jerk; this behavior is consistent with its role as the active detachment step. The Strawberry$\rightarrow$Basket transfer exhibited intermediate timing but noticeably larger variability, suggesting that fruit transport after detachment was less constrained and more sensitive to local geometric conditions. These motion characteristics are consistent with the representative greenhouse trajectories shown in Figure~\ref{fig:greenhouse_ppo_traj}, where the PPO controller produces smooth sequential Home$\rightarrow$Target$\rightarrow$Bucket motions across different target configurations. At the task level, the system achieved a reaching success rate of 96.6\%, a grasp-and-pull success rate of 91.3\%, and an overall harvesting success rate of 84.3\%. The relatively small drop from reaching to grasp-and-pull indicates that the perception-to-reach pipeline and the PPO-guided manipulation remained reliable during target engagement. The larger drop from grasp-and-pull to final harvesting success suggests that the main limitation was not target approach, but post-contact fruit handling and retention, which was likely influenced by the use of the SG-b-h gripper, a soft general-purpose gripper rather than a strawberry-specific harvesting end-effector. Figure~\ref{fig:harvest_demo} shows representative snapshots of the harvesting process in the greenhouse.

\begin{table*}[t]
\centering
\small
\setlength{\tabcolsep}{4.5pt}
\renewcommand{\arraystretch}{1.12}
\caption{Greenhouse motion performance of the PPO-based harvesting controller. Values are reported as mean $\pm$ standard deviation. The greenhouse experiments involved 130 trials and 281 target strawberries. Here, H $\rightarrow$ S denotes Home to target Strawberry, and S $\rightarrow$ B denotes Strawberry to Basket.}
\label{tab:greenhouse_motion_ppo}
\begin{tabular}{lccccc}
\toprule
\textbf{Primitive} &
\makecell{\textbf{Duration}\\\textbf{(s)}} &
\makecell{\textbf{Trajectory}\\\textbf{length (m)}} &
\makecell{\textbf{Mean}\\\textbf{speed (m/s)}} &
\makecell{\textbf{RMS}\\\textbf{jerk (m/s$^3$)}} &
\makecell{\textbf{Peak}\\\textbf{jerk (m/s$^3$)}} \\
\midrule
H $\rightarrow$ S & $4.95 \pm 0.35$ & $0.98 \pm 0.23$ & $0.20 \pm 0.06$ & $617.5 \pm 61.2$   & $5548.4 \pm 525.6$ \\
Pull              & $2.22 \pm 0.75$ & $0.25 \pm 0.14$ & $0.53 \pm 0.24$ & $1180.4 \pm 674.5$ & $5500.2 \pm 623.6$ \\
S $\rightarrow$ B & $4.17 \pm 1.04$ & $1.23 \pm 0.25$ & $0.32 \pm 0.23$ & $879.0 \pm 1360.7$ & $5267.1 \pm 1222.2$ \\
\bottomrule
\end{tabular}
\end{table*}

Reported segment durations reflect only active motion execution time. Fixed waiting and dwell periods in the harvesting routine, such as detection waiting at the home pose, post-motion dwell intervals, and gripper pause times, were outside the logged segment interval and were therefore not included in the reported motion metrics.

Eventually, this study presents a comprehensive and practical robotic system for strawberry harvesting that effectively connects perception with task execution. It integrates a robust vision model with a simulation-trained DRL controller within a ROS-based system. This integration is often missing in conventional robotic harvesting architectures. The proposed vision model significantly outperformed the baseline YOLO26-seg, demonstrating impressive resilience under the cluttered, variable conditions of commercial greenhouse environments. Additionally, our perception module was specifically designed for manipulation-oriented 3D localization. This approach ensured that the generated targets were not only visually accurate but also kinematically feasible for the robotic arm. A core advantage of this framework lies in simulation-based reaching policy learning. By utilizing PPO within Isaac Sim/Lab, a high-fidelity simulation platform, the reaching controller was trained under randomized conditions and scalable rollouts. This approach eliminated the need for exhaustive physical trials, significantly reducing hardware dependency and development costs while outperforming traditional IK in generating smooth, target-conditioned trajectories within constrained agricultural workspaces. Ultimately, this work distinguishes itself from prior studies by delivering an end-to-end system that is simultaneously resource-efficient and reliable. Rather than relying on a platform-specific refinement process, the proposed methodology offers a scalable blueprint for autonomous harvesting, as summarized in the comparative analysis of state-of-the-art systems (Table~\ref{tab:system_level_comparison}).

Despite these advancements, several limitations highlight the need for further investigation. The current 3D target generation relies on segmented centroids and local depth values rather than full 6D fruit pose or peduncle geometry. While sufficient for the pull-based harvesting routine utilized here, more complex detachment strategies may require orientation-sensitive interaction and explicit grasp affordance. Additionally, the target-selection logic operates sequentially without active perception or occlusion-aware viewpoint adjustment, limiting task-level planning in highly dense environments. Furthermore, the reliance on calibrated frame transformations with empirical offsets introduces a potential sensitivity to hardware reconfigurations or varying greenhouse layouts. 

To move from a proof-of-concept toward a commercially viable platform, future work should therefore focus on two major directions. The first is richer target representation. Extending the current fruit-center target to include peduncle localization, fruit orientation, or grasp affordance prediction would make the system more suitable for precision interaction beyond the present pull-based routine. The second is broader autonomy and robustness. This includes more extensive cross-greenhouse validation, cultivar variation, active perception under occlusion, and tighter integration between perception uncertainty and action selection during harvesting.

\begin{table*}[t]
\small
\centering
\caption{System-level comparison of recent robotic harvesting frameworks with ours.}
\label{tab:system_level_comparison}
\setlength{\tabcolsep}{3.6pt}
\renewcommand{\arraystretch}{1.12}
\begin{tabular}{@{}llclccp{3.6cm}@{}}
\toprule
Study & Vision system & \begin{tabular}[c]{@{}c@{}}Vision\\enhance.\end{tabular} & Motion control & \begin{tabular}[c]{@{}c@{}}Learned\\reaching\end{tabular} & \begin{tabular}[c]{@{}c@{}}Sim-to-\\real\end{tabular} & Key limitations\\
\midrule
Huang et al.\cite{HUANG2026111616} & F2SA-YOLOv8 & \checkmark & HVSC & $\times$ & $\times$ & \begin{tabular}[t]{@{}l@{}}
• sensor-based calibration\\
• noise-sensitive motion\\
• singularity-prone
\end{tabular}\\

Wu et al.~\cite{WU2026111409} & YOLO11 & \checkmark & Visual servo & $\times$ & $\times$ & \begin{tabular}[t]{@{}l@{}}
• visibility \& FoV\\\hspace{1.2em}constrained\\
• unstable motion\\
• singularity-prone
\end{tabular}\\

Arad et al.~\cite{arad2020development} & Color-thresholding & $\times$ & IK-based Moveit & $\times$ & \checkmark & \begin{tabular}[t]{@{}l@{}}
• antedated vision system\\
• unstable motion\\
• singularity-prone
\end{tabular}\\

Bhat et al.~\cite{bhat2026vision} & YOLO11-seg & $\times$ & IK-based Moveit & $\times$ & \checkmark & \begin{tabular}[t]{@{}l@{}}
• pretrained vision system\\
• unstable motion\\
• singularity-prone
\end{tabular}\\

\textbf{Ours} & \begin{tabular}[c]{@{}l@{}}HRAttnEdge-\\YOLO26-seg\end{tabular} & \checkmark & DRL-based PPO & \checkmark & \checkmark & \begin{tabular}[t]{@{}l@{}}
• fruit-center targets\\
• use of commercial\\\hspace{1.2em}end-effector
\end{tabular}\\
\bottomrule
\end{tabular}
\end{table*}

\section{Conclusion}
\label{sec:conclusion}

A unified perception-to-action framework is presented in this study that integrates RGB-D fruit localization, simulation-trained PPO-based reaching, and ROS-based execution for greenhouse harvesting. At the perception level, the proposed HRAttnEdge-YOLO26-seg model improved instance-level fruit localization under cluttered and occluded conditions, yielding reliable targets for fruit manipulation. At the control level, the simulation-trained PPO provides a scalable alternative to hardware-intensive trial-and-error, enabling robust target-conditioned reaching while reducing the development costs and risks. Deployment on a UR10e platform demonstrated that combining specialized RGB-D perception with simulation-based learning yields a practical and effective pipeline that advances beyond conventional systems that rely on general-purpose perception and planner-driven control.

Despite these contributions, this study has several limitations. The current system operates on smoothed fruit-center targets derived from segmentation and depth alignment without explicitly modeling peduncle geometry, grasp affordances, or full 6D pose, which may be required for more complex interaction strategies. Camera-to-robot mapping depends on calibrated transformations with empirical offsets, which potentially limits portability across setups. In addition, the harvesting policy is sequential and reactive, lacking active viewpoint adaptation and occlusion-aware planning under dense canopies. These limitations outline clear directions for developing a more autonomous and generalizable, field-ready system.

\section*{Acknowledgments}
This research is partially supported by the United States Department of Agriculture (USDA)'s National Institute of Food and Agriculture (NIFA) Research Capacity Fund Hatch Program: TEX09954 (Accession No. 7002248) and Research Capacity Fund Multistate Hatch Program: TEX0-1-9916 (Accession No. 7008389). This publication is also supported by the Texas A\&M University Division of Research Targeted Proposal Teams (TPT) funding program and Texas A\&M AgriLife Research. Any opinions, findings, conclusions, or recommendations expressed in this publication are those of the authors and should not be construed to represent any official USDA or U.S. Government determination or policy. 

\section*{Credit authorship contribution statement}
Al Bashir.: Conceptualization, Methodology, Formal analysis, Data curation, Validation, Writing original – review \& editing, 
draft, Visualization. Shao-Yang Chang.: Conceptualization, Methodology, Validation, Writing – review \& editing. Partho Ghose.: Writing – review \& editing. Prem Raj.: Writing – review \& editing. Chen-Kang Huang.: Writing – review \& editing. Azlan Zahid: Supervision, Funding acquisition, Writing – review \& editing. 

\section*{Conflict of interest}
The authors declare no competing interests.

 \bibliographystyle{elsarticle-num}
 \bibliography{source}

@misc{yolo26_ultralytics,
  author = {Glenn Jocher and Jing Qiu},
  title = {Ultralytics YOLO26},
  version = {26.0.0},
  year = {2026},
  url = {https://github.com/ultralytics/ultralytics},
  orcid = {0000-0001-5950-6979, 0000-0003-3783-7069},
  license = {AGPL-3.0}
}

@misc{yolo11_ultralytics,
  author = {Glenn Jocher and Jing Qiu},
  title = {Ultralytics YOLO11},
  version = {11.0.0},
  year = {2024},
  url = {https://github.com/ultralytics/ultralytics},
  orcid = {0000-0001-5950-6979, 0000-0003-3783-7069},
  license = {AGPL-3.0}
}

@misc{yolov8_ultralytics,
  author = {Glenn Jocher and Ayush Chaurasia and Jing Qiu},
  title = {Ultralytics YOLOv8},
  version = {8.0.0},
  year = {2023},
  url = {https://github.com/ultralytics/ultralytics},
  orcid = {0000-0001-5950-6979, 0000-0002-7603-6750, 0000-0003-3783-7069},
  license = {AGPL-3.0}
}

@article{Stemmler2023,
   author = {Henry Stemmler and Eva Marie Meemken},
   doi = {10.1016/j.foodpol.2023.102443},
   issn = {03069192},
   journal = {Food Policy},
   month = {5},
   publisher = {Elsevier Ltd},
   title = {Greenhouse farming and employment: Evidence from Ecuador},
   volume = {117},
   year = {2023},
}

@article{charlton2019can,
  title={Can wages rise quickly enough to keep workers in the fields?},
  author={Charlton, Diane and Taylor, J Edward and Vougioukas, Stavros and Rutledge, Zachariah},
  journal={Choices},
  volume={34},
  number={2},
  pages={1--7},
  year={2019},
  publisher={JSTOR}
}

@article{schimmelpfennig2018crop,
  title={Crop production costs, profits, and ecosystem stewardship with precision agriculture},
  author={Schimmelpfennig, David},
  journal={Journal of Agricultural and Applied Economics},
  volume={50},
  number={1},
  pages={81--103},
  year={2018},
  publisher={Cambridge University Press}
}

@article{pankova2020economic,
  title={Economic aspects of precision agriculture systems},
  author={P{\'a}nkov{\'a}, Ludmila and Aulov{\'a}, Renata and Jarol{\'\i}mek, Jan},
  journal={AGRIS on-line Papers in Economics and Informatics},
  volume={12},
  number={3},
  pages={59--67},
  year={2020},
  publisher={Czech University of Life Sciences Prague, Faculty of Economics and Management}
}

@article{getahun2024application,
  title={Application of precision agriculture technologies for sustainable crop production and environmental sustainability: A systematic review},
  author={Getahun, Sewnet and Kefale, Habtamu and Gelaye, Yohannes},
  journal={The Scientific World Journal},
  volume={2024},
  number={1},
  pages={2126734},
  year={2024},
  publisher={Wiley Online Library}
}

@article{win2025labor,
  title={Labor shortages and farmer adaptation strategies},
  author={Win, Myat Thida and Rutledge, Zachariah and Maredia, Mywish K},
  journal={Applied Economic Perspectives and Policy},
  volume={47},
  number={3},
  pages={896--913},
  year={2025},
  publisher={Wiley Online Library}
}

@incollection{pandey2026robotics,
  title={Robotics and Automation in Modern Agriculture: Revolutionizing Harvesting and Processing},
  author={Pandey, Shreya and Kaushik, Kashish and Tewatia, Anjali and Quraishi, Suhail Javed},
  booktitle={Precision and Intelligence in Agriculture: Advanced Technologies for Sustainable Farming},
  pages={153--186},
  year={2026},
  publisher={IGI Global Scientific Publishing}
}

@article{HUANG2026111616,
title = {A robotic harvesting system for occluded cucumbers using F2SA-YOLOv8 and HVSC},
journal = {Computers and Electronics in Agriculture},
volume = {246},
pages = {111616},
year = {2026},
issn = {0168-1699},
doi = {10.1016/j.compag.2026.111616},
author = {Zhi-Hong Huang and Chun-Ta Chen and Naoki Ikegaya and Tien‑Li Chang and Kun‑Cheng Ke and Yi-Chia Chen}
}

@article{jia2020detection,
  title={Detection and segmentation of overlapped fruits based on optimized mask R-CNN application in apple harvesting robot},
  author={Jia, Weikuan and Tian, Yuyu and Luo, Rong and Zhang, Zhonghua and Lian, Jian and Zheng, Yuanjie},
  journal={Computers and Electronics in Agriculture},
  volume={172},
  pages={105380},
  year={2020},
  publisher={Elsevier}
}

@article{wang2025research, 
  title={Research progress and development trend of visual detection methods for selective fruit harvesting robots},
  author={Wang, Wenbo and Li, Chenshuo and Xi, Yidan and Gu, Jinan and Zhang, Xinzhou and Zhou, Man and Peng, Yuchun},
  journal={Agronomy},
  volume={15},
  number={8},
  pages={1926},
  year={2025},
  publisher={MDPI}
}

@article{espinoza2023analysis, 
  title={Analysis of fruit images with deep learning: A systematic literature review and future directions},
  author={Espinoza, Sebastian and Aguilera, Cristhian and Rojas, Luis and Campos, Pedro G},
  journal={IEEE Access},
  volume={12},
  pages={3837--3859},
  year={2023},
  publisher={IEEE}
}

@article{tang2023optimization,
  title={Optimization strategies of fruit detection to overcome the challenge of unstructured background in field orchard environment: A review},
  author={Tang, Yunchao and Qiu, Jiajun and Zhang, Yunqi and Wu, Dongxiao and Cao, Yuhong and Zhao, Kexin and Zhu, Lixue},
  journal={Precision Agriculture},
  volume={24},
  number={4},
  pages={1183--1219},
  year={2023},
  publisher={Springer}
}

@article{noda2025robot,
  title={Robot Simulation on Agri-Field Point Cloud With Centimeter Resolution},
  author={Noda, Shintaro and Kogoshi, Masayuki and Iijima, Wataru},
  journal={IEEE Access},
  volume={13},
  pages={14404--14416},
  year={2025},
  publisher={IEEE}
}

@article{catala2024harnessing,
  title={Harnessing digital twins for agriculture 5.0: a comparative analysis of 3D point cloud tools},
  author={Catala-Roman, Paula and Navarro, Enrique A and Segura-Garcia, Jaume and Garcia-Pineda, Miguel},
  journal={Applied Sciences},
  volume={14},
  number={5},
  pages={1709},
  year={2024},
  publisher={MDPI}
}

@article{calzada2025review,
  title={A review on inverse kinematics, control and planning for robotic manipulators with and without obstacles via deep neural networks},
  author={Calzada-Garcia, Ana and Victores, Juan G and Naranjo-Campos, Francisco J and Balaguer, Carlos},
  journal={Algorithms},
  volume={18},
  number={1},
  pages={23},
  year={2025},
  publisher={MDPI}
}

@article{al2024enhancing, 
  title={Enhancing robotic autonomy: a review and case study of traditional and deep learning approaches to inverse kinematics},
  author={Al-Sharif, Omar A and Abbass, Nourhan A and Hanafi, Ahmed M and Elnady, Abdelrady O},
  journal={Int J Eng Appl Sci Oct 6 Univ},
  volume={1},
  number={1},
  pages={1--8},
  year={2024}
}

@article{rajendran2024towards,
  title={Towards autonomous selective harvesting: A review of robot perception, robot design, motion planning and control},
  author={Rajendran, Vishnu and Debnath, Bappaditya and Mghames, Sariah and Mandil, Willow and Parsa, Soran and Parsons, Simon and Ghalamzan-E, Amir},
  journal={Journal of Field Robotics},
  volume={41},
  number={7},
  pages={2247--2279},
  year={2024},
  publisher={Wiley Online Library}
}

@article{liu2024review,
  title={Review on motion planning of robotic manipulator in dynamic environments},
  author={Liu, Jie and Yap, Hwa Jen and Khairuddin, Anis Salwa Mohd},
  journal={Journal of Sensors},
  volume={2024},
  number={1},
  pages={5969512},
  year={2024},
  publisher={Wiley Online Library}
}

@article{soleymanzadeh2026towards,
  title={Towards Generalist Neural Motion Planners for Robotic Manipulators: Challenges and Opportunities},
  author={Soleymanzadeh, Davood and Lopez-Sanchez, Ivan and Su, Hao and Li, Yunzhu and Liang, Xiao and Zheng, Minghui},
  journal={IEEE Transactions on Automation Science and Engineering},
  year={2026},
  publisher={IEEE}
}

@article{schulman2017proximal,
  title={Proximal policy optimization algorithms},
  author={Schulman, John and Wolski, Filip and Dhariwal, Prafulla and Radford, Alec and Klimov, Oleg},
  journal={arXiv preprint arXiv:1707.06347},
  year={2017}
}

@article{elumalai2025proximal,
  title={A proximal policy optimization based deep reinforcement learning framework for tracking control of a flexible robotic manipulator},
  author={Elumalai, Vinodh Kumar and others},
  journal={Results in Engineering},
  volume={25},
  pages={104178},
  year={2025},
  publisher={Elsevier}
}

@inproceedings{silveira2025simulation,
  title={A Simulation Pipeline to Facilitate Real-World Robotic Reinforcement Learning Applications},
  author={Silveira, Jefferson and Marshall, Joshua A and Givigi, Sidney N},
  booktitle={2025 IEEE International systems Conference (SysCon)},
  pages={1--8},
  year={2025},
  organization={IEEE}
}

@article{HERNANDEZ2025110161, 
title = {Reconfigurable agricultural robotics: Control strategies, communication, and applications},
journal = {Computers and Electronics in Agriculture},
volume = {234},
pages = {110161},
year = {2025},
issn = {0168-1699},
doi = {10.1016/j.compag.2025.110161},
author = {Henry Alberto Hernández and Iván Fernando Mondragón and Sergio Ramiro González and Luis Fernando Pedraza}
}

@article{malis20022,
  title={2 1/2 D visual servoing},
  author={Malis, Ezio and Chaumette, Francois and Boudet, Sylvie},
  journal={IEEE Transactions on Robotics and Automation},
  volume={15},
  number={2},
  pages={238--250},
  year={2002},
  publisher={IEEE}
}

@article{chaumette2006visual,
  title={Visual servo control. I. Basic approaches},
  author={Chaumette, Fran{\c{c}}ois and Hutchinson, Seth},
  journal={IEEE robotics \& automation magazine},
  volume={13},
  number={4},
  pages={82--90},
  year={2006},
  publisher={IEEE}
}

@ARTICLE{1321162,
  author={Chesi, G. and Vicino, A.},
  journal={IEEE Transactions on Robotics}, 
  title={Visual servoing for large camera displacements}, 
  year={2004},
  volume={20},
  number={4},
  pages={724-735},
  doi={10.1109/TRO.2004.829465}
}

@article{levine2018learning,
  title={Learning hand-eye coordination for robotic grasping with deep learning and large-scale data collection},
  author={Levine, Sergey and Pastor, Peter and Krizhevsky, Alex and Ibarz, Julian and Quillen, Deirdre},
  journal={The International journal of robotics research},
  volume={37},
  number={4-5},
  pages={421--436},
  year={2018},
  publisher={SAGE Publications Sage UK: London, England}
}

@article{arad2020development,
  title={Development of a sweet pepper harvesting robot},
  author={Arad, Boaz and Balendonck, Jos and Barth, Ruud and Ben-Shahar, Ohad and Edan, Yael and Hellstr{\"o}m, Thomas and Hemming, Jochen and Kurtser, Polina and Ringdahl, Ola and Tielen, Toon and others},
  journal={Journal of field robotics},
  volume={37},
  number={6},
  pages={1027--1039},
  year={2020},
  publisher={Wiley Online Library}
}

@article{WU2026111409,
title = {Robot visual servo based on lightweight YOLO11-SMMA for Camellia oleifera fruits harvesting},
journal = {Computers and Electronics in Agriculture},
volume = {243},
pages = {111409},
year = {2026},
issn = {0168-1699},
doi = {10.1016/j.compag.2026.111409},
author = {Shangshang Wu and Haifei Chen and Yuyan Zhang and Kai Liao and Lijun Li}
}

@article{kurup2026comparative,
  title={Comparative Analysis Between Proximal Policy Optimization and Its Applications in High-Complexity Sequential Decision-Making},
  author={Kurup, Archa and Sumesh, Devika and Saju, Linete Lisa and Suresh, Sonal},
  journal={INTERNATIONAL JOURNAL OF ENGINEERING DEVELOPMENT AND RESEARCH},
  volume={14},
  number={1},
  pages={504--510},
  year={2026},
  publisher={IJEDR (www. ijedr. org)}
}

@article{bhat2026vision,
  title={Vision-guided Autonomous Dual-arm Extraction Robot for Bell Pepper Harvesting},
  author={Bhat, Kshitij Madhav and Gao, Tom and Mathur, Abhishek and Satishkumar, Rohit and Yandun, Francisco and Bauer, Dominik and Pollard, Nancy},
  journal={arXiv preprint arXiv:2603.13987},
  year={2026}
}

@article{liu2022digital,
  title={A digital twin-based sim-to-real transfer for deep reinforcement learning-enabled industrial robot grasping},
  author={Liu, Yongkui and Xu, He and Liu, Ding and Wang, Lihui},
  journal={Robotics and Computer-Integrated Manufacturing},
  volume={78},
  pages={102365},
  year={2022},
  doi={10.1016/j.rcim.2022.102365},
  publisher={Elsevier}
}

@incollection{HUAPENG2001321,
title = {Inverse kinematics analysis of a parallel redundant manipulator by means of differential evolution},
editor = {Eiji Arai and Tatsuo Arai and Masaharu Takano},
booktitle = {Human Friendly Mechatronics},
publisher = {Elsevier Science},
address = {Amsterdam},
pages = {321-326},
year = {2001},
isbn = {978-0-444-50649-8},
doi = {10.1016/B978-044450649-8/50054-1},
author = {Wu Huapeng and Heikki Handroos}
}

@article{arulkumaran2017deep,
  title={Deep reinforcement learning: A brief survey},
  author={Arulkumaran, Kai and Deisenroth, Marc Peter and Brundage, Miles and Bharath, Anil Anthony},
  journal={IEEE signal processing magazine},
  volume={34},
  number={6},
  pages={26--38},
  year={2017},
  publisher={IEEE}
}

@ARTICLE{11219194,
  author={Bashir, Al and Wang, Yibin and Ojo, Mike O. and Zahid, Azlan},
  journal={IEEE Transactions on AgriFood Electronics}, 
  title={A Vision System for Occluded Cutting Point Localization in Robotic Harvesting of Greenhouse Lettuce}, 
  year={2026},
  volume={4},
  number={1},
  pages={157-167},
  doi={10.1109/TAFE.2025.3621592}}

@article{schwarke2025rslrl,
  title={RSL-RL: A Learning Library for Robotics Research},
  author={Schwarke, Clemens and Mittal, Mayank and Rudin, Nikita and Hoeller, David and Hutter, Marco},
  journal={arXiv preprint arXiv:2509.10771},
  year={2025}
}

@article{PEREZBORRERO2020105736,
title = {A fast and accurate deep learning method for strawberry instance segmentation},
journal = {Computers and Electronics in Agriculture},
volume = {178},
pages = {105736},
year = {2020},
issn = {0168-1699},
doi = {10.1016/j.compag.2020.105736},
author = {Isaac Pérez-Borrero and Diego Marín-Santos and Manuel E. Gegúndez-Arias and Estefanía Cortés-Ancos}
}

@misc{Sekachev2020,
   author = {Boris Sekachev and Nikita Manovich and Maxim Zhiltsov and Andrey Zhavoronkov and Dmitry Kalinin and Ben Hoff and TOsmanov and Dmitry Kruchinin and Artyom Zankevich and DmitriySidnev and Maksim Markelov and Johannes222 and Mathis Chenuet and a-andre and telenachos and Aleksandr Melnikov and Jijoong Kim and Liron Ilouz and Nikita Glazov and Priya4607 and Rush Tehrani and Seungwon Jeong and Vladimir Skubriev and Sebastian Yonekura and vugia truong and zliang7 and lizhming and Tritin Truong},
   month = {8},
   publisher = {Zenodo},
   title = {opencv/cvat: v1.1.0},
   doi = {10.5281/zenodo.4009388},
   year = {2020},
}

@article{mittal2025isaac,
  title={Isaac lab: A gpu-accelerated simulation framework for multi-modal robot learning},
  author={Mittal, Mayank and Roth, Pascal and Tigue, James and Richard, Antoine and Zhang, Octi and Du, Peter and Serrano-Munoz, Antonio and Yao, Xinjie and Zurbr{\"u}gg, Ren{\'e} and Rudin, Nikita and others},
  journal={arXiv preprint arXiv:2511.04831},
  year={2025},
  doi={10.48550/arXiv.2511.04831}
}

@article{everingham2010pascal,
  title={The pascal visual object classes (voc) challenge},
  author={Everingham, Mark and Van Gool, Luc and Williams, Christopher KI and Winn, John and Zisserman, Andrew},
  journal={International journal of computer vision},
  volume={88},
  number={2},
  pages={303--338},
  year={2010},
  publisher={Springer},
  doi={10.1007/s11263-009-0275-4}
}

@inproceedings{lin2014microsoft, 
  title={Microsoft coco: Common objects in context},
  author={Lin, Tsung-Yi and Maire, Michael and Belongie, Serge and Hays, James and Perona, Pietro and Ramanan, Deva and Doll{\'a}r, Piotr and Zitnick, C Lawrence},
  booktitle={European conference on computer vision},
  pages={740--755},
  year={2014},
  organization={Springer},
  doi = {10.1007/978-3-319-10602-1_48}
}

@article{ramer1972iterative,
  title={An iterative procedure for the polygonal approximation of plane curves},
  author={Ramer, Urs},
  journal={Computer graphics and image processing},
  volume={1},
  number={3},
  pages={244--256},
  year={1972},
  publisher={Elsevier}
}

@article{douglas1973algorithms, 
  title={Algorithms for the reduction of the number of points required to represent a digitized line or its caricature},
  author={Douglas, David H and Peucker, Thomas K},
  journal={Cartographica: the international journal for geographic information and geovisualization},
  volume={10},
  number={2},
  pages={112--122},
  year={1973},
  publisher={University of Toronto Press}
}
\end{document}